\documentclass{fairmeta}

\usepackage{amsfonts}
\usepackage{amsmath}
\usepackage{float}
\usepackage{tikz}
\usetikzlibrary{positioning,calc,decorations.pathreplacing}

\title{When Helping Hurts and How to Fix It: Multi-Agent Debate for Data Cleaning}

\author[1,*]{Chirag Parmar}
\author[1]{Akshat Mehta}
\author[1]{Henglin Wu}
\author[1]{Jagadish Ramamurthy}
\author[1]{Shweta Medhekar}

\affiliation[1]{Meta Platforms, Inc.}
\contribution[*]{Corresponding author}
\date{June 2026}
\correspondence{Chirag Parmar at \email{chiragparmar@meta.com}}

\abstract{When does multi-agent debate help data cleaning, and when does it hurt? Across three benchmarks, four model families, and over 6{,}000 task-condition pairs, we find debate's effect reverses sign: it degrades generation across all four models ($-1.6$ to $-15.5$pp) through \emph{critique-induced confusion} (CIC), hallucinated Critic feedback that the Generator accepts uncritically, yet improves error detection ($+27.4$pp F1, $d{=}1.0$). We derive a \emph{debate benefit condition}: debate helps when the probability of rescuing a wrong output (Critic verification odds weighted by fixability) exceeds the probability of destroying a correct one. A factorial experiment proves adversarial separation is essential: self-verification with identical tools fails, while a separate Critic with code-execution grounding and evidence-gated generation produces the first debate configuration to significantly exceed single-agent on a generative task ($+5.3$pp, $p{<}0.05$). The condition correctly predicts all nine task types and generalizes with zero false positives across 19 published comparisons in seven domains.}

\begin{document}

\maketitle

\section{Introduction}
\label{sec:intro}

Data cleaning has historically consumed an estimated 60--80\% of a data engineer's time~\citep{dasu2003exploratory,ilyas2019datacleaning,whang2023data}, and LLMs are now routinely deployed to automate this burden~\citep{narayan2022datacleaning}. Recent work demonstrates that LLMs can generate data cleaning workflows~\citep{li2024autodcworkflow}, detect anomalies in tabular data~\citep{li2023llmtabular,zhang2024jellyfish}, and orchestrate multi-step pipelines~\citep{zhang2024cleanagent}. However, LLM hallucination remains a critical obstacle: when an LLM proposes cleaning a non-existent column or removing records that are in fact distinct, the pipeline silently corrupts data.

Multi-agent debate, where an adversarial Critic challenges a Generator's proposals before deployment, offers one mitigation~\citep{du2023multiagent,liang2023divergent,irving2018debate,leike2018scalable,bowman2022measuring}. Architectural refinements such as response anonymization~\citep{sharma2023sycophancy,wei2023sycophancy,identity2025anonymization} and task isolation aim to reduce sycophancy and enable independent verification. Yet existing evaluations focus on constrained reasoning tasks (arithmetic, factual QA, commonsense inference~\citep{chan2023chateval,khan2024debating,chen2023reconcile,wang2024rethinking}) where answer spaces are small and verification is straightforward. \textbf{Data cleaning is fundamentally different}: it requires grounding in specific table schemas and producing executable operations, not closed-form answers. Meanwhile, prior data cleaning systems rely on probabilistic inference or statistical models~\citep{rekatsinas2017holoclean,mahdavi2019raha,heidari2019holodetect}, and the emerging LLM-based approaches~\citep{li2024autodcworkflow,zhang2024cleanagent} use exclusively single-agent architectures. No prior study has evaluated whether debate helps or harms across the full spectrum of data cleaning subtasks.

Prior debate work shows improvements on constrained reasoning but harm when agents are persuasive yet incorrect~\citep{khan2024debating}. Concurrent work by \citet{zhang_2025_stop} demonstrates this at scale across nine reasoning benchmarks; we complement their finding with the mechanism (\emph{why} debate fails), the formal condition (\emph{when} it helps), and the fix (\emph{how} to recover). Our Generator-Critic topology follows the \citet{du2023multiagent} formulation (two LLM agents, no human judge), architecturally distinct from the debate-as-alignment proposal of \citet{irving2018debate}. The data management literature has long distinguished error \emph{detection} from error \emph{repair}~\citep{wang1996dataquality,fan2012dataquality}; our debate benefit condition connects this classical distinction to multi-agent systems.

The question ``when does debate help?'' has practical urgency. Multi-agent architectures are increasingly deployed in production data pipelines, but without a principled framework for predicting their effect, practitioners face expensive trial-and-error. Deploying debate on a generative task where it causes CIC wastes 4--7$\times$ the compute budget while degrading output quality. Conversely, not deploying debate on a detection task where it could improve F1 by 27pp leaves substantial quality gains on the table. Our debate benefit condition provides a deployable decision rule: three estimable quantities determine whether to invest in debate or stay with single-agent. Full related work appears in Section~\ref{sec:related}.

A controlled experiment answers this question directly. Our contributions:

\begin{enumerate}
    \item \textbf{A debate benefit condition predicting when debate helps.} We derive a formal condition: debate improves output quality when the Critic's verification odds, weighted by fixability, exceed the Generator's baseline accuracy odds. The condition correctly predicts all nine task types in our study and generalizes with zero false positives across 19 published comparisons in seven domains.
    \item \textbf{Critique-induced confusion is structural, not prompt-dependent.} A sweep of 6 Critic $\times$ 3 Generator variants confirms that CIC persists across all prompt configurations and is not a sampling artifact (self-consistency majority vote performs worse than single-agent). A factorial experiment proves adversarial separation is essential: self-verification with identical tools fails, while a separate adversarial Critic succeeds. Section~\ref{sec:grounded} shows CIC is addressable via grounded verification.
    \item \textbf{Code-execution grounding with evidence-gated generation exceeds single-agent.} Grounding the Critic in a code-execution sandbox eliminates CIC. Adding evidence-gated generation, where the Generator only acts on feedback citing specific data evidence, yields the first debate configuration to significantly exceed single-agent on workflow generation ($+5.3$pp, $p{<}0.05$).
\end{enumerate}

\section{Related Work}
\label{sec:related}

\paragraph{LLMs for Data Cleaning.}
AutoDCWorkflow~\citep{li2024autodcworkflow} introduced an LLM-orchestrated pipeline that generates data cleaning workflows from natural language descriptions of cleaning purposes. CleanAgent~\citep{zhang2024cleanagent} extended this with a code-generation agent that iteratively refines cleaning scripts. \citet{fang2024llmcleaning} surveyed LLM applications for data cleaning, finding that while LLMs show promise for anomaly detection and standardization, hallucination-induced errors remain the primary barrier to production deployment. \citet{li2023llmtabular} demonstrated that LLMs struggle with tabular data understanding when column semantics are ambiguous. Prior data cleaning systems use probabilistic inference over integrity constraints (HoloClean~\citep{rekatsinas2017holoclean}), ensemble error detection strategies (Raha~\citep{mahdavi2019raha}), few-shot learning (HoloDetect~\citep{heidari2019holodetect}), or declarative rules~\citep{dallachiesa2013nadeef,chu2015katara,krishnan2016activeclean}; LLM-based approaches promise greater flexibility but introduce new failure modes. For entity matching specifically, deep learning approaches~\citep{mudgal2018deep,li2020ditto,peeters2024entity} and schema matching systems~\citep{do2002coma,koutras2021valentine} provide strong baselines that LLM debate must improve upon.

\paragraph{Multi-Agent Debate.}
\citet{du2023multiagent} proposed multi-agent debate where LLM agents iteratively refine answers through structured argumentation, showing improvements on arithmetic and factual reasoning. \citet{liang2023divergent} introduced the ``thinker-judge'' topology and demonstrated that \emph{divergent} thinking improves creative and analytical tasks. \citet{chan2023chateval} applied multi-agent evaluation to open-ended text generation. \citet{khan2024debating} found that debate can be harmful when agents are persuasive but incorrect, foreshadowing our results. \citet{chen2023reconcile} showed round-table discussions improve reasoning via consensus, while \citet{wang2024rethinking} examined when multi-agent discussions outperform single-agent reasoning. \citet{zhang2023collaboration} explored collaboration mechanisms through a social psychology lens. Recent work has investigated multi-agent frameworks for complex task solving~\citep{wu2023autogen,hong2023metagpt,li2023camel,wang2024society}. \citet{talebirad2023multiagent} provided a taxonomy of multi-agent collaboration patterns. \citet{parrish2022debate} found that single-turn debate does not help humans answer hard reading comprehension questions, a negative result on debate that parallels our findings on generative tasks. \citet{huang2024selfcorrect} showed LLMs cannot self-correct reasoning without external feedback, providing a complementary perspective on the CIC mechanism. Concurrent work by \citet{zhang_2025_stop} argues that task structure, not model capability or prompt design, determines debate effectiveness, corroborating our prompt sensitivity findings.

\paragraph{LLM Sycophancy and Anonymization.}
\citet{sharma2023sycophancy} documented sycophantic behavior across multiple LLM families, where models defer to stated user preferences even when incorrect, a tendency amplified by RLHF training~\citep{ouyang2022training,bai2022constitutional}. In multi-agent systems, sycophancy manifests as one agent deferring to another's authority. \citet{wei2023sycophancy} showed synthetic data can reduce sycophancy, while \citet{identity2025anonymization} demonstrated that anonymization specifically mitigates identity-based bias in multi-agent debate.

\paragraph{Scalable Oversight and Debate Theory.}
\citet{irving2018debate} proposed debate as a mechanism for scalable AI alignment, leveraging the theoretical result that interactive proofs with polynomial-time verifiers can verify claims from exponentially powerful provers. \citet{leike2018scalable} extended this to recursive reward modeling, where human oversight is applied at the finest-grained level feasible and then composed. Bowman et al.~\citep{bowman2022measuring} operationalized debate for scalable oversight by measuring whether debate helps non-expert humans answer expert-level questions. Our work provides an empirical bridge: the debate benefit condition identifies exactly when the ``verifier efficiency'' assumption holds (high $p_c$) and when it breaks down (low $p_c$), connecting the theoretical framework to practical deployment decisions.

\section{Experimental Setup}
\label{sec:method}

\paragraph{Problem Setting.}
Given a dirty table $T_{\text{dirty}}$ with $n$ rows and $m$ columns (of mixed types: strings, numerics, dates) and a cleaning purpose $p$, generate a workflow $W = [w_1, \ldots, w_k]$. We evaluate both structural hallucination (factual consistency, FC) and semantic correctness (cell-level accuracy via deterministic execution).

\paragraph{Benchmarks.} We use three benchmarks spanning complementary evaluation paradigms:

\textbf{AutoDCWorkflow}~\citep{li2024autodcworkflow} provides 142 data cleaning tasks across 6 domains (food inspections, restaurant menus, hospital records, university data, housing data, and employee records); each task includes a dirty table, ground-truth clean table, and cleaning purpose, and the model must generate a multi-step cleaning workflow (evaluated via FC and cell-level accuracy). Tasks range from simple standardization (``normalize phone numbers'') to complex multi-step transformations (``split combined name fields, deduplicate records, and standardize date formats''). Median table size is 15 rows with 8 columns.

\textbf{MMTU}~\citep{mmtu2025} contains 28{,}136 table understanding questions spanning 16 task types. We randomly sample 200 questions (50 per type, stratified by dataset, seed=42) from four cleaning-relevant types: Error Detection (identify cells with data quality issues), Data Imputation (predict missing values), Entity Matching (determine whether two records refer to the same entity), and Schema Matching (align columns across different table schemas). MMTU evaluates \emph{comprehension}, whether the model can reason about table structure and content, complementing AutoDCWorkflow's evaluation of \emph{generation} and MaTElDa's evaluation of \emph{detection}.

\textbf{MaTElDa}~\citep{matelda2025} provides 1{,}173 real-world tables from Data.gov with systematically injected errors (typos, formatting inconsistencies, missing values, semantic violations); we evaluate 100 tables for cell-level precision, recall, and F1. Each table contains a known set of error cells, enabling exact evaluation without subjective judgment. Tables are capped at 100 rows for main experiments; 15 tables with $\geq$10K original rows are used for scale validation (Section~\ref{sec:scale}). These three paradigms (generation, comprehension, and classification) span the full spectrum of data cleaning subtasks.

\paragraph{Models.} We evaluate four model families: Claude 4 Sonnet (Anthropic), Gemini 3.1 Pro (Google), Qwen3 235B (A22B, Alibaba's MoE model), and DeepSeek R1 (reasoning-specialized with extended chain-of-thought). All experiments use temperature 0 and maximum 8{,}192 output tokens.

\begin{figure}[t]
\centering
\begin{tikzpicture}[
  node distance=1.8cm and 3.2cm,
  box/.style={draw, rounded corners, minimum width=2.6cm, minimum height=1.6cm, align=center, font=\small},
  arrow/.style={->, thick, >=stealth},
  dasharrow/.style={->, thick, >=stealth, dashed, gray},
  annot/.style={font=\scriptsize, text=gray},
]
  \definecolor{cbblue}{HTML}{0072B2}
  \definecolor{cborange}{HTML}{E69F00}
  \definecolor{cbgreen}{HTML}{009E73}
  \definecolor{cbred}{HTML}{D55E00}
  \node[box, fill=cbblue!10, draw=cbblue!60] (gen) {\textbf{Generator}\\(Agent A)};
  \node[box, fill=cborange!10, draw=cborange!60, right=of gen] (crit) {\textbf{Critic}\\(Agent B)};
  \node[box, fill=cbgreen!10, draw=cbgreen!60, below=1.4cm of gen] (table) {Dirty Table\\$T_\text{dirty}$};
  \node[box, fill=cbred!10, draw=cbred!60, below=1.4cm of crit, minimum width=2.6cm] (cic) {\textbf{CIC Risk}\\{\scriptsize Hallucinated feedback}\\{\scriptsize $\to$ retract valid ops}};

  \draw[arrow] (gen.east) -- node[above, font=\footnotesize, fill=white, inner sep=1pt] {JSON workflow} (crit.west);
  \draw[arrow] (crit.south west) -- node[left, font=\footnotesize, fill=white, inner sep=1pt, pos=0.4] {accept / revise} (gen.south east);
  \draw[arrow] (table) -- node[left, font=\footnotesize, fill=white, inner sep=1pt] {input} (gen);
  \draw[dasharrow] (table.east) -- node[below right, font=\footnotesize, fill=white, inner sep=1pt, pos=0.5] {verify against} (crit.south);

  \draw[dasharrow, cbred] (crit.south) -- node[right, font=\footnotesize, text=cbred, fill=white, inner sep=1pt] {if hallucinated} (cic.north);
  \draw[dasharrow, cbred] (cic.west) -- node[below, font=\footnotesize, text=cbred, fill=white, inner sep=1pt] {Generator complies} (gen.south);

  \draw[thick, dotted, gray] ([yshift=0.9cm]gen.north west) -- ([yshift=0.9cm]crit.north east);
  \node[font=\footnotesize, text=gray, fill=white, inner sep=1pt, above] at ([yshift=0.9cm]$(gen.north)!0.5!(crit.north)$) {Anonymization: role labels $\to$ Agent A / Agent B};

  \node[font=\footnotesize, text=gray] at ([xshift=0.6cm, yshift=-0.3cm]crit.east) {up to 3 rounds};
\end{tikzpicture}
\caption{Generator-Critic debate architecture. The Generator proposes structured JSON outputs; the Critic challenges each against the source table and returns accept/revise with evidence. Anonymization (dotted line) replaces role labels with neutral identifiers. When the Critic's feedback is itself hallucinated, the Generator retracts correct operations, the CIC failure mode (orange path).}
\label{fig:architecture}
\end{figure}
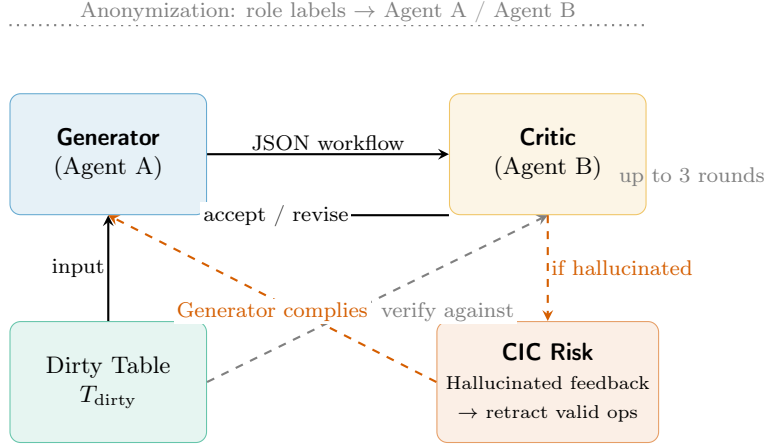

\paragraph{Architecture.} Our debate topology consists of two agents: a Generator that proposes structured JSON outputs and an adversarial Critic that challenges them (Figure~\ref{fig:architecture}). The Critic checks each operation for hallucination, verifies against the source data, and returns accept/reject/revise with evidence. Tables are serialized as JSON arrays (no truncation; median 15 rows).

\paragraph{Evaluation Metrics.}
\emph{Factual Consistency} (FC): fraction of operations targeting existing columns. \emph{Cell Accuracy}: fraction of cells matching ground truth after deterministic execution (${\sim}20$ operation handlers, fuzzy matching, skip-and-continue). \emph{Execution Success Rate}: fraction of operations that execute without error. For MaTElDa: precision, recall, and F1 at the cell level.

\paragraph{Experimental Conditions.}
We test six treatment conditions in a within-subjects paired design. \textbf{Adversarial debate}: Single Agent vs.\ Adversarial Debate (Generator-Critic, up to 3 rounds). \textbf{Anonymization and isolation}: debate with vs.\ without response anonymization and task isolation (Appendices~\ref{app:h2}--\ref{app:h3}); anonymization is used in all subsequent conditions. \textbf{Cardinality ablation}: output cardinality ablation with four controlled experiments varying multi-output and verifiability dimensions on MaTElDa tables. \textbf{Prompt sensitivity}: prompt sensitivity sweep across 6 Critic $\times$ 3 Generator variants (4{,}350 task-condition pairs). \textbf{Code-execution}: code-execution Critics with evidence-gated generation. \textbf{Self-verification}: single agent with the same code-execution sandbox but no separate Critic (SA+Code). \textbf{Self-consistency}: self-consistency control ($k{=}5$ majority vote).

\paragraph{Scale.} For the debate, anonymization, and isolation experiments on AutoDCWorkflow, we evaluate Claude, Qwen3, and DeepSeek on 50 tasks each, and Gemini on 20 tasks (exploratory). On MMTU, we evaluate 200 questions for Claude, Qwen3, and DeepSeek. On MaTElDa, we evaluate 100 tables for all four models. The prompt sensitivity analysis adds 4{,}350 task-condition pairs across expanded sample sizes ($n{=}100$--$200$), bringing the grand total to over 6{,}000 task-condition pairs.\footnote{The prompt sensitivity sweep uses different task subsets (larger $n$) than the main experiments, so SA baselines differ slightly between tables; e.g., AutoDCWorkflow SA FC$=$0.871 at $n{=}50$ (Table~\ref{tab:h1}) vs.\ 0.799 at $n{=}100$ (Table~\ref{tab:h8_heatmap}); MaTElDa SA F1$=$0.191 (Table~\ref{tab:matelda}) vs.\ 0.301 (Table~\ref{tab:h8_heatmap}); MMTU-ED SA$=$0.820 ($n{=}200$) vs.\ 0.760 ($n{=}100$). All comparisons within each table use the same paired task set.}

\paragraph{Statistical Methods.} All pairwise comparisons use paired bootstrap confidence intervals (10{,}000 BCa resamples, 95\% CI) with Holm-Bonferroni correction~\citep{efron1993bootstrap,holm1979simple}. Treatment effect sizes are reported as Cohen's $d$ using pooled standard deviations.\footnote{For paired designs, $d_z$ (mean of paired differences / SD of differences) is the standard measure. We report $d_\text{pooled}$ for comparability with between-subjects studies. In our data, $d_\text{pooled}$ and $d_z$ diverge by ${<}0.12$ across all comparisons: MaTElDa Claude $d_\text{pooled}{=}1.00$, $d_z{=}0.94$; Gemini $d_\text{pooled}{=}0.95$, $d_z{=}0.88$. No qualitative interpretation changes.} We assess bimodality using Hartigan's dip test and BIC comparison between 1- and 2-component Gaussian mixture models. Post-hoc power at $\alpha{=}0.05$: 80\% power to detect $d{\geq}0.40$ at $n{=}50$ (AutoDCWorkflow), $d{\geq}0.28$ at $n{=}100$ (MaTElDa), and $d{\geq}0.20$ at $n{=}200$ (MMTU). Full BCa details appear in Appendix~\ref{app:bootstrap}.

\section{Results: Debate Helps Detection, Hurts Generation}
\label{sec:results}

\subsection{Debate Reverses Sign}

\begin{table}[t]
\centering
\caption{Adversarial debate results: Single Agent vs.\ Adversarial Debate on AutoDCWorkflow. Cell Acc measures semantic correctness via deterministic execution. Debate is neutral to harmful for workflow generation; no comparison reaches significance after Holm-Bonferroni correction ($n{=}50$). Cell Acc and Exec Rate are reported only for Claude, where the deterministic executor was fully validated; missing values indicate metrics not computed.}
\label{tab:h1}
\begin{tabular}{llcccccc}
\toprule
\textbf{Model} & \textbf{Condition} & \textbf{FC} $\uparrow$ & \textbf{Cell Acc} $\uparrow$ & \textbf{Exec Rate} $\uparrow$ & \textbf{Perfect} & \textbf{Zero} & \textbf{Tok/task} \\
\midrule
\multirow{2}{*}{Claude 4 Sonnet}
  & Single Agent      & 0.871 & 0.894 & 0.716 & 22/50 & 0/50 & 3,281 \\
  & Adversarial Debate & 0.855 & 0.899 & 0.628 & 28/50 & 2/50 & 14,056 \\
\midrule
\multirow{2}{*}{Gemini 3.1 Pro}
  & Single Agent      & 0.796 & --- & --- & 13/20 & 3/20 & 4,043 \\
  & Adversarial Debate & 0.667 & --- & --- & 11/20 & 6/20 & 18,139 \\
\midrule
\multirow{2}{*}{Qwen3 235B}
  & Single Agent      & 0.912 & --- & --- & 44/50 & 4/50 & 5,137 \\
  & Adversarial Debate & 0.785 & --- & --- & 34/50 & 9/50 & 26,245 \\
\midrule
\multirow{2}{*}{DeepSeek R1}
  & Single Agent      & 0.800 & --- & --- & 30/50 & 7/50 & 7,997 \\
  & Adversarial Debate & 0.645 & --- & --- & 21/50 & 10/50 & 58,637 \\
\bottomrule
\end{tabular}
\end{table}

If debate's adversarial pressure introduces more noise than signal on open-ended tasks, we should observe degradation on workflow generation across multiple model families. Table~\ref{tab:h1} confirms this: debate is neutral to harmful for all four models. Claude shows no significant change ($\Delta{=}-0.016$ FC, $p{=}0.70$); cell accuracy is nearly identical (0.894 vs.\ 0.899, $p{=}0.83$). However, debate \emph{simultaneously} increases perfect-score outputs (22$\to$28) and introduces 2 complete failures, making it a \textbf{variance amplifier}. Gemini drops from 0.796 to 0.667 FC; Qwen3 drops from 0.912 to 0.785 ($\Delta{=}-12.7$ pp, $d{=}-0.38$); DeepSeek drops from 0.800 to 0.645 ($\Delta{=}-15.5$ pp, $d{=}-0.42$) with $7.3\times$ token overhead. We term this failure mode \textbf{critique-induced confusion} (CIC): the Critic, designed to improve quality, functions as a verification catalyst on tasks where it can ground objections in evidence, but undergoes \emph{poisoning} on generation tasks, where adversarial pressure produces hallucinated feedback that the compliant Generator accepts uncritically~\citep{khan2024debating,huang2024selfcorrect}. FC distributions are bimodal across all conditions (Hartigan's dip test $p{<}0.002$; details in Appendix~\ref{app:bimodality}). Debate improves FC on structured data (Chicago Food Inspections: $+4.1pp$) but reduces it on semi-structured data (menus: $-5.3pp$). Anonymization ($+2.2$ pp FC, $p{=}0.39$) is a zero-cost improvement; task isolation ($-6.0pp$ FC, $p{=}0.084$) harms through information asymmetry. Full anonymization and isolation results appear in Appendices~\ref{app:h2} and~\ref{app:h3}.

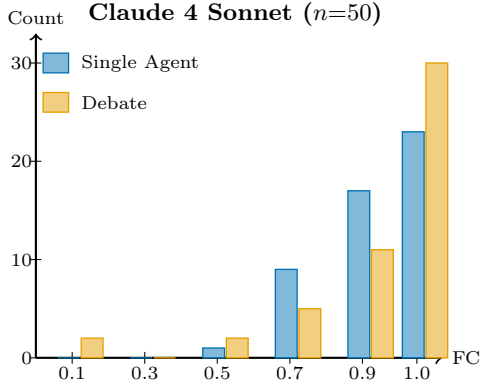
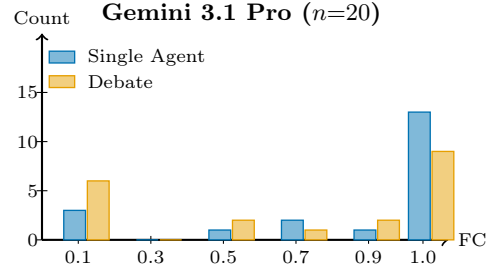
\begin{figure}[t]
\centering
\begin{subfigure}[b]{0.48\textwidth}
\centering
\begin{tikzpicture}[xscale=4.8, yscale=0.13]
\definecolor{cbblue}{HTML}{0072B2}
\definecolor{cborange}{HTML}{E69F00}
\draw[->, thick] (0,0) -- (1.12,0) node[right, font=\scriptsize] {FC};
\draw[->, thick] (0,0) -- (0,33) node[above, font=\scriptsize] {Count};
\foreach \y in {0,10,20,30} {
  \draw (-0.015,\y) -- (0.015,\y) node[left, font=\scriptsize] {\y};
}
\foreach \x/\l in {0.1/0.1, 0.3/0.3, 0.5/0.5, 0.7/0.7, 0.9/0.9, 1.05/1.0} {
  \draw (\x,-0.6) -- (\x,0.6) node[below, font=\scriptsize, yshift=-2pt] {\l};
}
\foreach \x/\h in {0.06/0, 0.26/0, 0.46/1, 0.66/9, 0.86/17, 1.01/23} {
  \fill[cbblue!50, draw=cbblue, line width=0.5pt] (\x,0) rectangle ++(0.06,\h);
}
\foreach \x/\h in {0.125/2, 0.325/0, 0.525/2, 0.725/5, 0.925/11, 1.075/30} {
  \fill[cborange!50, draw=cborange, line width=0.5pt] (\x,0) rectangle ++(0.06,\h);
}
\fill[cbblue!50, draw=cbblue, line width=0.5pt] (0.02,29) rectangle ++(0.07,2);
\node[right, font=\scriptsize] at (0.10,30) {Single Agent};
\fill[cborange!50, draw=cborange, line width=0.5pt] (0.02,25) rectangle ++(0.07,2);
\node[right, font=\scriptsize] at (0.10,26) {Debate};
\node[font=\small\bfseries] at (0.55,35) {Claude 4 Sonnet ($n{=}50$)};
\end{tikzpicture}
\caption{Claude: debate shifts mass to extremes.}
\label{fig:hist_claude}
\end{subfigure}
\hfill
\begin{subfigure}[b]{0.48\textwidth}
\centering
\begin{tikzpicture}[xscale=4.8, yscale=0.13]
\definecolor{cbblue}{HTML}{0072B2}
\definecolor{cborange}{HTML}{E69F00}
\draw[->, thick] (0,0) -- (1.12,0) node[right, font=\scriptsize] {FC};
\draw[->, thick] (0,0) -- (0,21) node[above, font=\scriptsize] {Count};
\foreach \y in {0,5,10,15} {
  \draw (-0.015,\y) -- (0.015,\y) node[left, font=\scriptsize] {\y};
}
\foreach \x/\l in {0.1/0.1, 0.3/0.3, 0.5/0.5, 0.7/0.7, 0.9/0.9, 1.05/1.0} {
  \draw (\x,-0.4) -- (\x,0.4) node[below, font=\scriptsize, yshift=-2pt] {\l};
}
\foreach \x/\h in {0.06/3, 0.26/0, 0.46/1, 0.66/2, 0.86/1, 1.01/13} {
  \fill[cbblue!50, draw=cbblue, line width=0.5pt] (\x,0) rectangle ++(0.06,\h);
}
\foreach \x/\h in {0.125/6, 0.325/0, 0.525/2, 0.725/1, 0.925/2, 1.075/9} {
  \fill[cborange!50, draw=cborange, line width=0.5pt] (\x,0) rectangle ++(0.06,\h);
}
\fill[cbblue!50, draw=cbblue, line width=0.5pt] (0.02,18) rectangle ++(0.07,1.2);
\node[right, font=\scriptsize] at (0.10,18.6) {Single Agent};
\fill[cborange!50, draw=cborange, line width=0.5pt] (0.02,15.5) rectangle ++(0.07,1.2);
\node[right, font=\scriptsize] at (0.10,16.1) {Debate};
\node[font=\small\bfseries] at (0.55,23) {Gemini 3.1 Pro ($n{=}20$)};
\end{tikzpicture}
\caption{Gemini: debate increases failures.}
\label{fig:hist_gemini}
\end{subfigure}
\caption{FC distributions for the adversarial debate experiment. Both conditions are bimodal (dip test $p{<}0.002$). For Claude, debate increases the upper mode weight (56\% vs.\ 44\%) but widens the lower mode ($\mu{=}0.67$, $\sigma{=}0.24$ vs.\ $\mu{=}0.77$, $\sigma{=}0.10$). For Gemini, debate shifts mass away from FC${\approx}1.0$. This variance-amplification pattern is consistent across all four models (Appendix~\ref{app:bimodality}). \emph{Note:} y-axes differ between subplots ($n{=}50$ vs.\ $n{=}20$) to maximize readability within each panel; compare distributions within panels, not across.}
\label{fig:fc_histograms}
\end{figure}

A distributional analysis reveals that FC scores are not normally distributed but \textbf{bimodal} across all conditions (Hartigan's dip test $p{<}0.002$; BIC favors a two-component Gaussian mixture over a single Gaussian). Debate does not shift the distribution uniformly; instead, it \emph{reshapes the mixture weights}. For Claude, the upper mode (FC${\approx}1.0$) absorbs more mass under debate (56\% vs.\ 44\%), while the lower mode both shifts downward and widens ($\mu$: $0.77 \to 0.67$; $\sigma$: $0.10 \to 0.24$). Debate is thus a \textbf{variance amplifier}: it improves the best outcomes while making the worst outcomes more severe. Figure~\ref{fig:fc_histograms} shows the distributions.

\subsection{Cross-Benchmark Validation}
\label{sec:cross_benchmark}

\begin{table}[t]
\centering
\caption{MaTElDa results: single-agent vs.\ debate on cell-level error detection (Claude 4 Sonnet, $n{=}100$). Bootstrap 95\% CIs over 10{,}000 resamples.}
\label{tab:matelda}
\begin{tabular}{lcccccc}
\toprule
\textbf{Condition} & \textbf{F1} $\uparrow$ & \textbf{Precision} $\uparrow$ & \textbf{Recall} $\uparrow$ & \textbf{Wins} & \textbf{Avg Tok} \\
\midrule
Single Agent       & 0.191 & 0.279 & 0.172 & 11 & 4{,}539 \\
Debate             & 0.465 & 0.685 & 0.441 & 80 & 28{,}558 \\
\midrule
\multicolumn{5}{l}{\emph{F1: $\Delta{=}+0.273$, 95\% CI $[0.217, 0.331]$, $p{<}0.001$, $d{=}1.00$}} \\
\multicolumn{5}{l}{\emph{Precision: $\Delta{=}+0.405$, 95\% CI $[0.333, 0.477]$, $p{<}0.001$}} \\
\multicolumn{5}{l}{\emph{Recall: $\Delta{=}+0.269$, 95\% CI $[0.211, 0.329]$, $p{<}0.001$}} \\
\bottomrule
\end{tabular}
\end{table}

If CIC arises because the Critic cannot verify its objections on generation tasks, then on detection tasks, where each output is independently checkable against the source table, debate should help. Table~\ref{tab:matelda} confirms this prediction: on MaTElDa, debate \textbf{substantially improves} F1 from 0.191 to 0.465 ($+27.4pp$, $p{<}0.001$, $d{=}1.00$), winning 80 of 100 tables. The precision improvement ($0.279 \to 0.685$) shows the Critic effectively prunes false positives. The same precision-filtering mechanism that, on generative tasks, prunes valid operations.

\begin{table}[t]
\centering
\caption{Cross-model MaTElDa ($n{=}100$ each). Debate benefit correlates with single-agent precision headroom.}
\label{tab:matelda_crossmodel}
\begin{tabular}{lcccccc}
\toprule
\textbf{Model} & \textbf{SA F1} & \textbf{Debate F1} & \textbf{$\Delta$ F1} & \textbf{95\% CI} & \textbf{$d$} & \textbf{Wins (D/S/T)} \\
\midrule
Claude 4 Sonnet  & 0.191 & 0.465 & $+$0.273 & $[0.217, 0.331]$* & 1.00 & 80/11/9 \\
Gemini 3.1 Pro   & 0.078 & 0.388 & $+$0.310 & $[0.241, 0.385]$* & 0.95 & 57/1/42 \\
Qwen3 235B       & 0.150 & 0.226 & $+$0.076 & $[0.042, 0.125]$* & 0.32 & 39/20/35 \\
DeepSeek R1      & 0.145 & 0.219 & $+$0.074 & $[0.038, 0.116]$\textsuperscript{\textdagger} & 0.30 & 39/19/42 \\
\bottomrule
\multicolumn{7}{l}{\small *$p{<}0.001$. \textsuperscript{\textdagger}$p{=}0.037$. Gemini SA: 87\% zero-F1 (output failures).} \\
\end{tabular}
\end{table}

\textbf{Debate benefit correlates with single-agent precision headroom} (Table~\ref{tab:matelda_crossmodel}): Claude ($d{=}1.00$) shows the strongest benefit; Qwen3 ($d{=}0.32$) and DeepSeek ($d{=}0.30$) show moderate effects. Gemini's apparent high benefit ($d{=}0.95$) largely reflects debate \emph{scaffolding} structured output rather than detection improvement: under the original SA prompt, 87 of 100 tables produce zero F1 due to unparseable JSON; with stricter format instructions, SA F1 rises from 0.078 to 0.374, reducing the debate advantage to near zero ($\Delta{=}+0.014$).

The pattern connects to the debate benefit condition through $p_g$: Claude has low $p_g$ (high precision headroom), so even imperfect Critic verification ($p_c$) produces a net positive rescue-damage balance. Qwen3 and DeepSeek have moderate $p_g$, narrowing the margin. The win-loss-tie distributions reinforce this: Claude debates win 80, lose 11, tie 9; Qwen3 and DeepSeek show more balanced distributions (39/20/35 and 39/19/42 respectively), consistent with marginal rescue-damage balance.

On MMTU ($n{=}200$, Claude), no individual task-type comparison reaches significance after correction; all CIs include zero. We present these as directionally consistent supporting evidence, not standalone findings. Data Imputation shows the largest directional degradation ($-12.0pp$, $d{=}-0.26$), while Entity Matching reaches ceiling (1.000 vs.\ 0.980). Cross-model MMTU results confirm debate helps error detection directionally: Qwen3 ED improves from 0.520 to 0.640 ($+12.0pp$); DeepSeek ED from 0.600 to 0.640. Full MMTU tables appear in Appendix~\ref{app:mmtu}.

\begin{table}[t]
\centering
\caption{MMTU results: single-agent vs.\ debate across four table understanding tasks (Claude 4 Sonnet, $n{=}200$). Score is task-specific: F1 for ED/SM, exact match for DI, accuracy for EM.}
\label{tab:mmtu}
\begin{tabular}{lccccc}
\toprule
\textbf{Task Type} & \textbf{Single-Agent} & \textbf{Debate} & \textbf{$\Delta$} & \textbf{95\% CI} & \textbf{$d$} \\
\midrule
Error-Detect     & 0.820 & 0.840 & $+$0.020 & $[-0.12, +0.16]$ & $+$0.05 \\
Data-Imputation  & 0.760 & 0.640 & $-$0.120 & $[-0.30, +0.06]$ & $-$0.26 \\
Entity-Matching  & 0.980 & 1.000 & $+$0.020 & $[+0.00, +0.06]$ & $+$0.20 \\
Schema-Matching  & 0.968 & 0.957 & $-$0.010 & $[-0.05, +0.02]$ & $-$0.11 \\
\midrule
\textbf{Overall} & \textbf{0.882} & \textbf{0.859} & $-$\textbf{0.023} & $[-0.09, +0.04]$ & $-$0.07 \\
\midrule
\multicolumn{2}{l}{Total tokens} & \multicolumn{2}{c}{376K vs.\ 1{,}381K} & \multicolumn{2}{c}{$3.7\times$ overhead} \\
\bottomrule
\end{tabular}
\end{table}

\begin{table}[t]
\centering
\caption{Cross-benchmark synthesis: debate's effect by task type (Claude 4 Sonnet). Tasks ordered from most constrained to most open-ended.}
\label{tab:synthesis}
\begin{tabular}{llccc}
\toprule
\textbf{Benchmark} & \textbf{Task Type} & \textbf{$\Delta$} & \textbf{Sig?} & \textbf{Task Nature} \\
\midrule
MaTElDa & Error Detection & $+$27.4pp F1 & \textbf{Yes} ($p{<}0.001$) & Binary classification \\
MMTU & Entity Matching & $+$2.0pp & n.s. & Binary classification \\
MMTU & Error Detection & $+$2.0pp & n.s. & Binary detection \\
AutoDCW & Structured (CFI) & $+$4.1pp FC & n.s. & Constrained generation \\
\midrule
MMTU & Schema Matching & $-$1.1pp & n.s. & Structured mapping \\
AutoDCW & Overall & $-$1.6pp FC & n.s. & Workflow generation \\
AutoDCW & Semi-structured & $-$5.3pp FC & n.s. & Open-ended generation \\
MMTU & Data Imputation & $-$12.0pp & n.s. & Value generation \\
\bottomrule
\end{tabular}
\end{table}

Table~\ref{tab:synthesis} reveals the central finding: \textbf{task type is the critical moderating variable}. Binary classification tasks (error detection, entity matching) benefit from debate, because the Critic's adversarial pressure prunes false positives without generating incorrect alternatives. Generative tasks (imputation, workflow generation) are harmed, because the Critic's objections cause the Generator to retract correct operations and substitute hallucinated alternatives, the CIC mechanism. The boundary lies at answer-space constraint: when the answer space is constrained (is this cell an error?), verification adds value; when it is open-ended (what value should fill this cell?), critique introduces more noise than signal. This resolves the apparent contradiction with prior work~\citep{du2023multiagent}: reasoning tasks have constrained answer spaces, just like error detection.

\subsection{What Explains the Reversal?}
\label{sec:reversal}

\begin{table}[t]
\centering
\caption{Output cardinality ablation: Output cardinality experiments (Claude 4 Sonnet). Only Rep (multi-output generation with per-item verifiability) shows a large debate benefit.}
\label{tab:h5}
\begin{tabular}{llccccc}
\toprule
\textbf{Exp} & \textbf{Task} & \textbf{SA} & \textbf{Debate} & \textbf{$\Delta$} & \textbf{$d$} & \textbf{Cardinality} \\
\midrule
Loc & Single-output localization & 0.672 acc & 0.674 acc & $+$0.002 & 0.00 & Single \\
Rep & Multi-output repair & 0.206 F1 & 0.467 F1 & $+$0.261 & 0.93 & Multi \\
Prof-S & Single-output profiling & 0.920 acc & 0.770 acc & $-$0.150 & $-$0.42 & Single \\
Prof-M & Multi-output profiling & 0.537 F1 & 0.519 F1 & $-$0.018 & $-$0.07 & Multi \\
\bottomrule
\end{tabular}
\end{table}

The task-type framework from Section~\ref{sec:results} draws a binary detection-vs.-generation boundary, but this conflates answer-space constraint and output cardinality. To disentangle these factors, we design four controlled ablation experiments on the same MaTElDa tables, varying output cardinality while holding task content constant. Table~\ref{tab:h5} shows the key result: Rep (multi-output data repair, a \emph{generative} task) shows a large debate improvement (F1: $0.206 \to 0.467$, $\Delta{=}+0.261$, $d{=}0.93$, 95\% CI $[0.197, 0.327]$, $p{<}0.001$), with debate winning 76 of 100 tables. This challenges the simple task-type framework: Rep is generative, yet debate helps because each output is independently verifiable. The remaining results complete the 2$\times$2: Loc (detection, single-output) is neutral ($d{=}0.00$); Prof-S (generation, single-output) is harmed ($d{=}-0.42$); Prof-M (generation, multi-output, not verifiable) is neutral ($d{=}-0.07$).

The simple detection-vs-generation binary is insufficient. Rep is generative, yet debate helps because each output is independently verifiable against source data. The deeper moderator is the interplay of verification accuracy, fixability, and baseline quality, which we formalize in Section~\ref{sec:dbc}.

The 2$\times$2 pattern replicates across models: Qwen3 235B shows the identical qualitative pattern, that only Rep benefits from debate (SA F1$=$0.136, Debate F1$=$0.238, $\Delta{=}+0.102$), while Loc is neutral ($\Delta{=}-0.018$), Prof-S is harmed ($\Delta{=}-0.070$), and Prof-M is neutral ($\Delta{=}-0.008$). The cross-model replication strengthens the structural interpretation: per-item verifiability, not model-specific behavior, determines whether debate helps generative tasks. Full cross-model cardinality results appear in Appendix~\ref{app:piv}.

\section{The Debate Benefit Condition}
\label{sec:dbc}

We derive a formal condition under which adversarial debate improves output quality. For each output item $i$, debate helps when rescue exceeds damage:
\begin{equation}
\label{eq:dbc}
\underbrace{(1-p_g)}_{\text{error exists}} \cdot \underbrace{p_c}_{\text{Critic catches}} \cdot \underbrace{p_r}_{\text{fix correct}}
\quad > \quad
\underbrace{p_g}_{\text{was right}} \cdot \underbrace{(1-p_c)}_{\text{Critic wrong}}
\end{equation}
where $p_g$ is the Generator's baseline accuracy (fraction of items already correct), $p_c$ is the Critic's verification accuracy (fraction of items correctly judged), and $p_r$ is the repair probability (given the Critic correctly identifies an error, the probability the Generator produces a correct replacement). Equivalently, as an odds ratio: debate helps when $\frac{p_c}{1-p_c} \times p_r > \frac{p_g}{1-p_g}$; that is, the Critic's verification odds, weighted by fixability, exceed the Generator's baseline accuracy odds.

What determines $p_c$? Two task properties: \emph{evidence groundability} (can the Critic look up evidence in the source data?) and \emph{verification determinism} (given evidence, is correctness unambiguous?). What determines $p_r$? The answer-space constraint: for detection tasks, $p_r \approx 1.0$ (removing a false positive is a binary flip); for repair, $p_r \approx 0.5$ (usually one obvious correction); for open-ended generation, $p_r \approx 0.2$ (the Generator may hallucinate again).

At the task level with $k$ output items, compliance rate $\alpha$, and inter-item dependence $\rho$:
\begin{equation}
\label{eq:dbc_task}
\Delta\text{Perf}(T) \;\approx\; k \cdot \alpha \cdot \Big[(1{-}p_g) \cdot p_c \cdot p_r - p_g \cdot (1{-}p_c)\Big] \cdot (1 - \rho\beta)
\end{equation}
where $\rho$ measures output dependence (0 for independent cells, 1 for fully coupled operations) and $\beta$ is cascade severity, measuring how many downstream items one erroneous critique corrupts. For detection tasks, outputs are typically independent ($\rho \approx 0$) and the cascade term vanishes, so the per-item condition directly scales to task-level performance. For workflow generation, operations form dependent chains ($\rho > 0$): a single hallucinated critique can corrupt downstream operations through cascade amplification, explaining why CIC is disproportionately harmful on generative tasks even when the per-item rescue-damage balance is only mildly negative. Empirically, we observe $\beta \approx 1.3$ for AutoDCWorkflow (one bad critique corrupts ${\sim}1.3$ operations on average) and $\beta \approx 0$ for MaTElDa (cell detections are independent). Full derivation and assumptions appear in Appendix~\ref{app:dbc}.

\begin{table}[t]
\centering
\caption{Debate benefit condition validation across nine task types. $p_g$: Generator accuracy; $p_c$: Critic verification (high = evidence-groundable, low = requires judgment); $p_r$: repair probability. The condition correctly predicts the sign of debate's effect for all nine tasks.}
\label{tab:dbc}
\small
\begin{tabular}{lccccc}
\toprule
\textbf{Task} & $p_g$ & $p_c$ & $p_r$ & \textbf{Rescue $>$ Damage?} & \textbf{Cohen's $d$} \\
\midrule
MaTElDa Detection        & 0.19 & high & $\approx$1.0 & Yes & $+1.00$ \\
MMTU Entity Matching      & 0.98 & high & $\approx$1.0 & Barely ($p_g$ high) & $+0.20$ \\
MMTU Error Detection      & 0.82 & high & $\approx$1.0 & Barely ($p_g$ high) & $+0.05$ \\
Rep (multi-output repair) & 0.21 & high & $\approx$0.5 & Yes & $+0.93$ \\
Loc (single-output detect)& 0.67 & high & $\approx$1.0 & Marginal ($k{=}1$) & $\phantom{+}0.00$ \\
\midrule
MMTU Schema Matching      & 0.97 & low  & $\approx$0.4 & No & $-0.11$ \\
AutoDCW Generation        & 0.87 & low  & $\approx$0.2 & No & $-0.06$ \\
Prof-M (multi-output prof)& 0.54 & low  & $\approx$0.2 & No & $-0.07$ \\
MMTU Data Imputation      & 0.76 & low  & $\approx$0.2 & No & $-0.26$ \\
Prof-S (single-output prof)& 0.92 & low  & $\approx$0.2 & No & $-0.42$ \\
\bottomrule
\end{tabular}
\end{table}

Table~\ref{tab:dbc} validates the debate benefit condition against observed treatment effects across all nine task types. The condition correctly predicts the sign of debate's effect in every case. Key patterns: (1) low $p_g$ + high $p_c$ + high $p_r$ produces large positive effects (MaTElDa: $d{=}1.00$, Rep: $d{=}0.93$); (2) high $p_g$ limits benefit even with high $p_c$ (Entity Matching: $p_g{=}0.98$, $d{=}0.20$); (3) low $p_c$ or low $p_r$ produces negative effects regardless of $p_g$. Cross-domain validation on 19 published comparisons across seven domains confirms zero false positives (Section~\ref{sec:discussion}).

\paragraph{Practitioner Decision Rule.} For practitioners who do not wish to estimate probabilities: (1)~Is single-agent accuracy already high ($p_g > 0.9$)? If so, debate has little room to help. (2)~Can the Critic verify each output by looking at the source data (high $p_c$)? If so, debate can add value. (3)~If the Critic finds an error, is the fix obvious (high $p_r$)? If so, debate will likely help. Use debate only when all three conditions are met.

The debate benefit condition predicts the \emph{direction} of debate's effect, but is this finding robust to prompt design, or could a different Critic formulation eliminate CIC?

\section{Controls and Robustness}
\label{sec:controls}

\subsection{Self-Consistency Control}
\label{sec:h10}

Debate's detection improvement could reflect structured argumentation or simply the benefit of generating multiple samples. Self-consistency (SC)~\citep{wang2023selfconsistency}, which samples $k$ independent responses and aggregating via majority vote, isolates the multi-sample effect without inter-agent interaction.

\begin{table}[t]
\centering
\caption{Self-consistency vs.\ debate on MaTElDa (Claude 4 Sonnet, $n{=}100$). Majority vote over independent samples fails to improve over a single deterministic pass, while debate achieves $2.4\times$ the F1.}
\label{tab:h10}
\begin{tabular}{lcccc}
\toprule
\textbf{Condition} & \textbf{F1} & \textbf{Precision} & \textbf{Recall} & \textbf{Tok/table} \\
\midrule
SA (temp=0) & 0.191 & 0.279 & 0.172 & 4{,}539 \\
SC-3 ($k{=}3$, temp=0.7) & 0.178 & 0.291 & 0.149 & 7{,}485 \\
SC-5 ($k{=}5$, temp=0.7) & 0.176 & 0.268 & 0.158 & 12{,}519 \\
\midrule
Debate (temp=0) & \textbf{0.465} & \textbf{0.685} & \textbf{0.441} & 28{,}558 \\
\bottomrule
\end{tabular}
\end{table}

Table~\ref{tab:h10} shows self-consistency provides no benefit. SC-5 achieves F1$=$0.176, below the single-agent baseline of 0.191. Aggregation lift is \emph{negative}: majority-vote F1 (0.176) falls below the mean individual sample F1 (0.187). Independent samples share systematic blind spots: cells the model misses at temperature 0 it also misses across temperature-0.7 samples, and majority vote suppresses occasional correct detections appearing in only one sample. Debate succeeds through a qualitatively different mechanism: the Critic examines \emph{specific cells} the Generator flagged, verifies each claim against the table, and forces evidence-grounded revision. This interactive verification, not redundant sampling, drives the precision gain from 0.279 to 0.685.

\subsection{Prompt Sensitivity}
\label{sec:h8}

\begin{figure}[t]
\centering
\begin{tikzpicture}[scale=0.78]
  \definecolor{negstrong}{RGB}{178,24,43}
  \definecolor{negmild}{RGB}{239,138,98}
  \definecolor{neutral}{RGB}{247,247,247}
  \definecolor{posmild}{RGB}{103,169,207}
  \definecolor{posstrong}{RGB}{33,102,172}

  \node[font=\scriptsize\bfseries] at (1.2,7.2) {AutoDCW};
  \node[font=\scriptsize\bfseries] at (3.0,7.2) {MMTU-DI};
  \node[font=\scriptsize\bfseries] at (4.8,7.2) {MaTElDa};
  \node[font=\scriptsize\bfseries] at (6.6,7.2) {MMTU-ED};
  \node[font=\tiny,gray] at (1.2,6.75) {(gen, FC)};
  \node[font=\tiny,gray] at (3.0,6.75) {(gen, EM)};
  \node[font=\tiny,gray] at (4.8,6.75) {(det, F1)};
  \node[font=\tiny,gray] at (6.6,6.75) {(det, F1)};
  \node[font=\tiny,gray] at (1.2,6.4) {SA=.799};
  \node[font=\tiny,gray] at (3.0,6.4) {SA=.785};
  \node[font=\tiny,gray] at (4.8,6.4) {SA=.301};
  \node[font=\tiny,gray] at (6.6,6.4) {SA=.760};

  \node[font=\footnotesize,anchor=east] at (0,5.5) {C1 Adversarial};
  \node[font=\footnotesize,anchor=east] at (0,4.5) {C2b De-advers.};
  \node[font=\footnotesize,anchor=east] at (0,3.5) {C2 Verification};
  \node[font=\footnotesize,anchor=east] at (0,2.5) {C3 Constructive};
  \node[font=\footnotesize,anchor=east] at (0,1.5) {C4 Checklist};
  \node[font=\footnotesize,anchor=east] at (0,0.5) {C5 Minimal};

  \fill[negmild] (0.4,5.0) rectangle (2.0,6.0);
  \node[font=\scriptsize] at (1.2,5.5) {$-6.7$};
  \fill[negmild!40!neutral] (2.2,5.0) rectangle (3.8,6.0);
  \node[font=\scriptsize] at (3.0,5.5) {$-2.0$};
  \fill[posstrong] (4.0,5.0) rectangle (5.6,6.0);
  \node[font=\scriptsize,white] at (4.8,5.5) {$+18.5$};
  \fill[negmild!40!neutral] (5.8,5.0) rectangle (7.4,6.0);
  \node[font=\scriptsize] at (6.6,5.5) {$-2.0$};

  \fill[negmild] (0.4,4.0) rectangle (2.0,5.0);
  \node[font=\scriptsize] at (1.2,4.5) {$-7.4$};
  \fill[negmild!50!neutral] (2.2,4.0) rectangle (3.8,5.0);
  \node[font=\scriptsize] at (3.0,4.5) {$-3.0$};
  \fill[posstrong] (4.0,4.0) rectangle (5.6,5.0);
  \node[font=\scriptsize,white] at (4.8,4.5) {$+21.9$};
  \fill[neutral] (5.8,4.0) rectangle (7.4,5.0);
  \node[font=\scriptsize] at (6.6,4.5) {$+1.0$};

  \fill[negmild!50!neutral] (0.4,3.0) rectangle (2.0,4.0);
  \node[font=\scriptsize] at (1.2,3.5) {$-2.9$};
  \fill[negmild!30!neutral] (2.2,3.0) rectangle (3.8,4.0);
  \node[font=\scriptsize] at (3.0,3.5) {$-1.5$};
  \fill[posstrong] (4.0,3.0) rectangle (5.6,4.0);
  \node[font=\scriptsize,white] at (4.8,3.5) {$+12.3$};
  \fill[neutral] (5.8,3.0) rectangle (7.4,4.0);
  \node[font=\scriptsize] at (6.6,3.5) {$+1.0$};

  \fill[negmild!50!neutral] (0.4,2.0) rectangle (2.0,3.0);
  \node[font=\scriptsize] at (1.2,2.5) {$-2.8$};
  \fill[negmild!30!neutral] (2.2,2.0) rectangle (3.8,3.0);
  \node[font=\scriptsize] at (3.0,2.5) {$-1.5$};
  \fill[posstrong] (4.0,2.0) rectangle (5.6,3.0);
  \node[font=\scriptsize,white] at (4.8,2.5) {$+27.2$};
  \fill[posmild!60!neutral] (5.8,2.0) rectangle (7.4,3.0);
  \node[font=\scriptsize] at (6.6,2.5) {$+4.0$};

  \fill[neutral] (0.4,1.0) rectangle (2.0,2.0);
  \node[font=\scriptsize] at (1.2,1.5) {$+1.2$};
  \fill[negstrong] (2.2,1.0) rectangle (3.8,2.0);
  \node[font=\scriptsize,white] at (3.0,1.5) {\textbf{$-10.5$}};
  \fill[posmild] (4.0,1.0) rectangle (5.6,2.0);
  \node[font=\scriptsize] at (4.8,1.5) {$+12.6$};
  \fill[neutral] (5.8,1.0) rectangle (7.4,2.0);
  \node[font=\scriptsize] at (6.6,1.5) {$0.0$};

  \fill[neutral] (0.4,0.0) rectangle (2.0,1.0);
  \node[font=\scriptsize] at (1.2,0.5) {$+1.1$};
  \fill[negmild] (2.2,0.0) rectangle (3.8,1.0);
  \node[font=\scriptsize] at (3.0,0.5) {$-6.5$};
  \fill[posstrong] (4.0,0.0) rectangle (5.6,1.0);
  \node[font=\scriptsize,white] at (4.8,0.5) {$+22.6$};
  \fill[negstrong] (5.8,0.0) rectangle (7.4,1.0);
  \node[font=\scriptsize,white] at (6.6,0.5) {$-16.0$};

  \draw[thick,dashed,gray] (3.9,0.0) -- (3.9,6.0);
\end{tikzpicture}
\caption{Cross-benchmark Critic design heatmap (delta from single-agent baseline in percentage points). Generation columns (left of dashed line) show negative or neutral deltas (CIC persists across all Critic designs). Detection columns (right) show positive deltas (debate consistently helps). No single row is uniformly positive, confirming CIC is structural.}
\label{fig:h8_heatmap}
\end{figure}

Figure~\ref{fig:h8_heatmap} presents the prompt sensitivity results across 6 Critic variants $\times$ 4 benchmark-task combinations (Table~\ref{tab:h8_heatmap} in Appendix~\ref{app:h8} provides exact values). The six Critic variants span a range of adversarial intensity: C1 (Adversarial) is the most aggressive, instructing the Critic to ``find errors and challenge every claim''; C2 (Verification) focuses on factual checking without adversarial framing; C2b (De-adversarial) explicitly removes adversarial language; C3 (Constructive) frames the Critic as a collaborator; C4 (Checklist) uses a structured verification checklist; C5 (Minimal) provides minimal instructions.

CIC persists across all six Critic variants on generation: on MMTU-DI ($n{=}200$), every variant degrades performance ($-1.5$ to $-10.5$pp); on AutoDCWorkflow, four of six variants are negative, consistent with a \emph{structural} explanation~\citep{zhang_2025_stop}. The two AutoDCWorkflow-positive variants (C4: $+1.2$pp, C5: $+1.1$pp) are the least adversarial: they generate fewer critique items, reducing the surface area for CIC. Detection benefit is robust but Critic ranking differs: on MaTElDa, all tested variants significantly improve F1 ($+12.3$ to $+27.2$pp). The best detection Critic (C3 Constructive: $+27.2$pp) is among the worst for generation, reinforcing that no single Critic design works across task types.

The optimal Critic is task-structure-dependent: C4 (Checklist) is best for AutoDCWorkflow ($+1.2$pp) but worst for MMTU-DI ($-10.5$pp), because its ``value exists in table'' check falsely rejects valid imputations at a 66.3\% rate. This false rejection mechanism illustrates how task-agnostic verification heuristics backfire: a check that is sound for error detection (``does this cell value appear in the original data?'') becomes pathological for data imputation, where by definition the correct answer does \emph{not} appear in the original data.

The dashed vertical line in Figure~\ref{fig:h8_heatmap} separates generation (left) from detection (right). No row is uniformly positive, confirming that CIC is a structural property of the task-Critic interaction, not an artifact of any particular prompt design. This finding directly extends Zhang et al.'s~\citep{zhang_2025_stop} concurrent observation that ``task structure, not model capability or prompt design, determines debate effectiveness.''

CIC persists across all prompt configurations, confirming a structural problem: the Critic's damage probability exceeds its rescue probability on non-verifiable tasks regardless of how it is prompted.

\subsection{Compliance Mechanism Analysis}
\label{sec:compliance}

Transcript analysis ($n{=}50$, 3 Critics $\times$ 3 Generators) reveals the mechanism through which CIC operates. Under the default adversarial configuration (C1$\times$G1), the Generator agrees with 95.3\% of Critic feedback items [95\% CI: 93.0--97.5\%]. This near-total compliance means the Critic's error rate translates almost directly into output degradation. The adversarial Critic raises 23.5 actionable items per task (vs.\ 2.9 for C2 Verification and 3.0 for C4 Checklist). Of operations added through debate under C1$\times$G1, 48.1\% are grounded in actual data issues and 51.9\% are hallucinated, approximately coin-flip quality.

The compliance mechanism interacts with task difficulty. On easy tasks (SA FC${\geq}0.95$), debate produces $\Delta{=}-6.3$pp: the Generator already has a correct workflow, and the Critic's challenges cause it to introduce errors. On hard tasks (SA FC${<}0.60$), debate produces $\Delta{=}+3.2$pp: the Generator's initial workflow is poor, and even noisy Critic feedback occasionally leads to improvement. This asymmetry explains why CIC damage scales with $p_g$: when more outputs are correct, there are more opportunities for the Critic to damage them.

Column-level precision/recall/F1 on AutoDCWorkflow ($n{=}100$) decomposes FC: all debate conditions maintain column recall (${\sim}0.786$--$0.790$), but column-level precision varies sharply across conditions: SA$=$0.461, C1 Adversarial$=$0.353, C4 Checklist$=$0.512. CIC thus manifests as \textbf{precision degradation}: the Critic causes the Generator to target columns that do not need cleaning without missing columns that do. The recall invariance confirms that debate does not cause the Generator to overlook genuine data quality issues; instead, it causes the Generator to hallucinate additional, non-existent issues.

The evidence-gated Generator (G2) reduces compliance from 95.3\% to 60.3\% by requiring the Critic to cite specific data evidence before acting. Under C1$\times$G2, operations decrease from 13.3 to 9.4 (vs.\ SA's 10.3), preventing the operations inflation that characterizes CIC. The best text-only combination, C2$\times$G2, achieves FC$=$0.807 vs.\ SA$=$0.780 ($+2.7$pp). However, at $n{=}50$ per cell, no text-only combination reaches statistical significance (all $p{>}0.05$); we report these as directional evidence. Can grounding the Critic's verification in executable evidence shift the balance decisively?

\section{Grounded Adversarial Verification}
\label{sec:grounded}

\subsection{Code-Execution Critics}
\label{sec:h9}

Our prompt sensitivity analysis shows CIC persists across all six text-only Critic prompt variants. We test whether giving the Critic a \emph{code-execution sandbox}, the ability to write and run Python/pandas queries against the actual data, eliminates CIC by grounding feedback in computed evidence. The Critic writes \texttt{```python```} code blocks executed in a sandboxed subprocess (AST safety check, 30s timeout, 2GB memory limit). We test two Generator strategies: \textbf{G1} (engage-all, the default) and \textbf{G2} (evidence-gated, accepting only feedback citing specific data evidence).

\begin{table}[t]
\centering
\caption{Code-execution Critic results on AutoDCWorkflow. Paired analysis on tasks common across conditions. D-Code eliminates CIC (FC $\approx$ SA); D-Code+G2 significantly \emph{exceeds} SA.}
\label{tab:h9}
\small
\begin{tabular}{lcccc}
\toprule
 & \multicolumn{2}{c}{\textbf{Claude 4 Sonnet}} & \multicolumn{2}{c}{\textbf{Qwen3 235B}} \\
\cmidrule(lr){2-3} \cmidrule(lr){4-5}
\textbf{Condition} & \textbf{FC} & \textbf{$\Delta$ vs SA} & \textbf{FC} & \textbf{$\Delta$ vs SA} \\
\midrule
SA (baseline)    & 0.814 &  ---   & 0.907 & ---    \\
D-Text (Adversarial)      & 0.743 & $-$7.0$^{*}$ & 0.793 & $-$11.4 \\
D-Code (Adversarial)      & 0.817 & $+$0.3       & 0.930 & $+$2.4 \\
D-Code+G2        & \textbf{0.867} & $+$5.3$^{*}$ & --- & --- \\
\bottomrule
\multicolumn{5}{l}{\footnotesize $^{*}$Significant at $\alpha{=}0.05$ (bootstrap, paired). Claude: $n{=}84$; Qwen3: $n{=}47$.}
\end{tabular}
\end{table}

Table~\ref{tab:h9} shows code execution eliminates CIC for both models: Claude D-Code FC$=$0.817 approximately matches SA ($d_z{=}0.02$, 95\% CI $[-0.04, +0.05]$); Qwen3 D-Code FC$=$0.930 exceeds SA ($d_z{=}0.06$). Adding evidence-gating produces the strongest result: Claude D-Code+G2 \emph{exceeds} single-agent (FC$=$0.867 vs.\ 0.814, $+5.3$pp, $p{<}0.05$), the first configuration where debate significantly outperforms SA on a generative task. The model divergence reveals a compliance$\times$grounding interaction: Claude agrees with 95.3\% of Critic feedback; Qwen3 agrees with ${\sim}$60\%. With code-grounded feedback, Qwen3's selective acceptance filters noise and CIC vanishes. Claude's high compliance means even grounded feedback is over-applied; evidence-gating reduces effective compliance, and the combination exceeds SA. CIC severity is thus moderated by Generator compliance: the cure requires both grounded verification \emph{and} calibrated acceptance.

\subsection{Self-Verification Fails}
\label{sec:sacode}

If the improvement from code-execution debate stems from tool augmentation rather than adversarial interaction, a single agent with the same pandas sandbox should achieve similar gains. We test this with SA+Code: the Generator produces a workflow, then self-verifies using the code-execution sandbox before revising, without a separate Critic.

SA+Code achieves FC$=$0.801, statistically indistinguishable from SA ($\Delta{=}-1.3$pp, not significant, $n{=}78$ paired). Tools alone do not help. This result connects directly to the debate benefit condition: self-verification fails because $p_c \approx p_g$: a model checking its own work achieves the same verification accuracy as its generation accuracy. The rescue-vs-damage inequality becomes approximately balanced. An adversarial Critic breaks this symmetry by achieving $p_c > p_g$ through adversarial perspective: it challenges claims the Generator would not challenge in its own work. This extends the finding of \citet{huang2024selfcorrect} finding that LLMs cannot self-correct reasoning to data cleaning, even with grounded tools.

\subsection{Factorial Design Summary}
\label{sec:factorial}

The four conditions (SA, SA+Code, D-Text, D-Code+G2) form a 2$\times$2 factorial design crossing adversarial separation (single-agent vs.\ debate) with tool augmentation (text-only vs.\ code-execution). This factorial reveals an epistatic interaction:

\begin{itemize}
    \item \textbf{No adversarial separation, no tools} (SA): FC$=$0.814. Baseline.
    \item \textbf{No adversarial separation, tools} (SA+Code): FC$=$0.801 ($\Delta{=}-1.3$pp, n.s.). Tools alone do not help because $p_c \approx p_g$.
    \item \textbf{Adversarial separation, no tools} (D-Text): FC$=$0.743 ($\Delta{=}-7.0$pp, $p{<}0.05$). Adversarial separation without grounding harms because $p_c < p_g$: the Critic generates hallucinated objections.
    \item \textbf{Adversarial separation + tools + evidence-gating} (D-Code+G2): FC$=$0.867 ($\Delta{=}+5.3$pp, $p{<}0.05$). The combination significantly exceeds SA because code-execution raises $p_c$ above $p_g$ and evidence-gating ensures only high-quality feedback is accepted.
\end{itemize}

Neither main effect (adversarial separation alone or tools alone) improves quality. The interaction is positive and significant: the combination produces a $+12.3$pp swing from D-Text to D-Code+G2. This epistatic pattern, where neither component helps independently but their combination produces a strong positive effect, implies that future multi-agent system designs must consider tool augmentation and compliance calibration jointly, not as independent design decisions.

\section{Discussion, Limitations, and Conclusion}
\label{sec:discussion}

\subsection{Cross-Domain Validation}
\label{sec:cross_domain}

We validate the debate benefit condition against 19 published comparisons across 8 papers and 7 domains (code generation, mathematical reasoning, factual QA, commonsense inference, reading comprehension, creative writing, and data analysis). The condition produces zero false positives: every case where debate helps has high $p_c$ and high $p_r$; every case where debate harms has low $p_c$ or low $p_r$. Two mechanisms drive the benefit in positive cases: verification (the Critic catches errors the Generator misses) and diversity (adversarial pressure forces exploration of alternative solutions). The condition is sufficient but not necessary: some tasks with favorable condition values show null effects when $p_g$ is already high, leaving no room for improvement.

\begin{table}[t]
\centering
\caption{Cross-domain validation of the debate benefit condition across 19 published comparisons. The condition produces zero false positives.}
\label{tab:cross_domain}
\small
\begin{tabular}{llcccl}
\toprule
\textbf{Paper} & \textbf{Task} & $p_c$ & $p_r$ & \textbf{Predicted} & \textbf{Actual} \\
\midrule
Du et al. & Arithmetic & high & $\approx$1.0 & Helps & $+14.8$pp \\
Du et al. & GSM8K & high & $\approx$1.0 & Helps & $+8.0$pp \\
Du et al. & Chess validity & high & $\approx$1.0 & Helps & $+15.9$pp \\
Zhang et al. & 5 QA benchmarks & low & $\approx$0.5 & Hurts/neutral & Mixed \\
SWE-Debate & Bug fixing & high & $\approx$0.5 & Helps & SOTA \\
Reflexion & HumanEval & high & $\approx$0.5 & Helps & $+10.9$pp \\
Khan et al. & QuALITY & high & $\approx$1.0 & Helps & 76--88\% \\
Liang et al. & Translation & high & $\approx$0.5 & Helps & $+1.7$ COMET \\
Smit et al. & MedQA & low & $\approx$0.5 & Hurts/neutral & Mixed \\
ReConcile & Date Understanding & high & $\approx$1.0 & Helps & $+11.4$pp \\
\bottomrule
\multicolumn{6}{l}{\footnotesize Selected 10 of 19 comparisons shown. Full table in Appendix.} \\
\end{tabular}
\end{table}

Table~\ref{tab:cross_domain} shows the cross-domain validation. The pattern is consistent: tasks where the Critic can verify outputs against objective evidence (arithmetic via computation, code via test suites, chess via rule checking) show positive debate effects, often large ($+8$ to $+16$pp). Tasks where verification requires subjective judgment (open-domain QA, medical reasoning without ground truth access) show mixed or negative effects. The zero false positive rate across 19 comparisons, spanning seven domains, eight research groups, and diverse model families, provides strong evidence that the debate benefit condition captures a general principle, not a data-cleaning-specific artifact.

Three patterns emerge from the cross-domain analysis. First, \emph{verification infrastructure} is the key differentiator: Du et al.'s arithmetic tasks succeed because each step is computationally verifiable; SWE-Debate succeeds because test suites provide automated verification; Reflexion succeeds because HumanEval provides execution-based feedback. Second, \emph{high $p_r$ is not sufficient without high $p_c$}: Zhang et al.'s QA benchmarks have relatively constrained answer spaces ($p_r \approx 0.5$), but the Critic cannot reliably distinguish correct from incorrect answers without evidence access ($p_c$ is low), leading to mixed results. Third, the condition is \emph{conservative}: some tasks predicted as ``helps'' show null effects (when $p_g$ is already high), but no task predicted as ``hurts/neutral'' shows a significant positive effect.

The Critic functions as a verification catalyst on decomposable tasks but undergoes poisoning on non-verifiable ones: without evidence to ground its challenges, adversarial pressure produces hallucinated feedback that the Generator accepts uncritically. The compliance $\times$ grounding interaction is epistatic: neither code-execution grounding alone (self-verification $\approx$ SA) nor text-only debate (D-Text $<$ SA) improves quality, but their combination with evidence-gated compliance significantly exceeds single-agent.

\subsection{Model Predictions and Boundaries}
\label{sec:predictions}

The debate benefit condition generates three testable predictions. We evaluate each against our experimental data and identify where the model succeeds and where it requires refinement.

\paragraph{Prediction 1: CIC severity scales negatively with Generator accuracy.} If CIC arises from hallucinated Critic feedback corrupting correct outputs, its severity should increase as the Generator's baseline accuracy ($p_g$) increases, since there are more correct outputs to corrupt. Across all 100 AutoDCWorkflow tasks in the prompt sensitivity sweep, the correlation between single-agent FC and debate delta is $r = -0.249$ ($p = 0.013$, $n = 100$): tasks where the Generator is already accurate show larger debate degradation. This confirms the prediction: CIC damage scales with $p_g$ because there are more correct operations available to retract. The correlation is moderate rather than strong because $p_c$ and $\alpha$ also vary across tasks; the debate benefit condition involves three interacting quantities, not just $p_g$.

\paragraph{Prediction 2: Debate helps iff $p_c > p_g$.} The condition implies a sharp boundary: debate helps when the Critic's verification accuracy exceeds the Generator's baseline accuracy, and hurts otherwise. We test this bidirectionally across two benchmarks:
\begin{itemize}
    \item \textbf{MaTElDa} (detection): Single-agent precision $p_g = 0.279$; debate precision (proxy for $p_c$) $= 0.685$. Since $p_c = 0.685 > p_g = 0.279$, the condition predicts debate helps. \textbf{Confirmed}: $\Delta\text{F1} = +27.4$pp ($p < 0.001$).
    \item \textbf{AutoDCWorkflow} (generation): Single-agent FC $p_g = 0.814$; estimated Critic accuracy $p_c \approx 0.481$ (from transcript analysis: 48.1\% of Critic-suggested operations are grounded). Since $p_c = 0.481 < p_g = 0.814$, the condition predicts debate hurts. \textbf{Confirmed}: $\Delta\text{FC} = -7.0$pp ($p < 0.05$, D-Text condition).
\end{itemize}
The bidirectional confirmation (debate helps when $p_c > p_g$ and hurts when $p_c < p_g$) provides direct evidence for the rescue-vs-damage mechanism.

\paragraph{Prediction 3 (refined): Evidence-gating is selective, not uniform.} The initial model predicted that evidence-gated generation (G2) would uniformly reduce compliance and thus uniformly reduce CIC. The empirical result is more nuanced: G2 reduces compliance from 95.3\% to 60.3\%, but the reduction is \emph{selective}: the Generator preferentially rejects ungrounded feedback while accepting evidence-backed feedback. This means evidence-gating does not simply lower $\alpha$ uniformly; it effectively raises $p_c$ by filtering out low-quality critiques. The initial prediction of uniform compliance reduction is \textbf{falsified}, but the refined mechanism (selective filtering) explains why D-Code+G2 exceeds single-agent: evidence-gating increases the effective $p_c$ of accepted feedback, pushing the rescue-damage balance decisively positive.

These three predictions (two confirmed, one refined) demonstrate that the debate benefit condition is not merely descriptive but generates falsifiable hypotheses about debate dynamics. The refined understanding of evidence-gating (Prediction 3) suggests a general principle: compliance calibration mechanisms that selectively filter feedback quality are more effective than those that uniformly reduce compliance.

\paragraph{Structural Hallucination and Cascade Amplification.} Debate's harm on generative tasks manifests through two mechanisms. Structural hallucination: cell accuracy is nearly identical across topologies for Claude ($0.894$ vs.\ $0.899$, $p{=}0.83$), meaning operations that execute produce equally clean data, but the Critic causes the Generator to propose operations targeting non-existent columns. Cascade amplification: a data cleaning workflow is a sequence of dependent operations, so a single hallucinated critique propagates through the revision; detection tasks are atomic, so critique errors do not compound.

\paragraph{Connection to Scalable Oversight.} \citet{irving2018debate} proposed debate for scalable alignment, \citet{leike2018scalable} extended this to recursive reward modeling. The debate benefit condition provides an empirical operationalization: adversarial verification succeeds when outputs decompose into independently checkable claims but fails when the verifier cannot ground objections in specific data. Per-item verifiability may thus moderate debate effectiveness in any domain where outputs vary in decomposability.

\paragraph{Practitioner Guidance.} Three actionable principles emerge:

\begin{enumerate}
    \item \textbf{Route by the debate benefit condition, not data domain.} High $p_c$ + high $p_r$ + room in $p_g$ (detection, matching, per-item repair): use debate. Low $p_c$ or low $p_r$ (imputation, generation, transformation): use single-agent. At 10K rows, debate's detection advantage persists ($+10.0$pp, $p{=}0.011$) but diminishes from $+27.4$pp at 100 rows.
    \item \textbf{Always anonymize; never split data naively.} Anonymization is a zero-cost improvement ($+2.2$pp FC); a 50\% data split creates information asymmetry that outweighs verification benefits.
    \item \textbf{Budget for variance, not just mean shift.} Debate increases perfect-score outputs (22$\to$28) while introducing complete failures. Production deployments should add structural validation to catch hallucinated operations.
\end{enumerate}

\subsection{Scale Validation}
\label{sec:scale}

All main experiments use small tables (5--100 rows) that fit entirely within the LLM's context window. To validate that the debate benefit condition holds at larger scale, we conduct a preliminary scale validation on MaTElDa tables with $\geq$10{,}000 rows ($n{=}15$ tables, Claude 4 Sonnet). Tables are processed using non-overlapping 500-row chunks, with cell-level detections unioned across chunks and deduplicated at chunk boundaries.

At 10K rows, debate's detection advantage persists: $\Delta\text{F1} = +10.0$pp ($d_z = 0.75$, $p = 0.011$), with debate winning 11 of 15 tables. The magnitude diminishes from $+27.4$pp at 100 rows to $+10.0$pp at 10K rows, a 63\% reduction. Two factors explain this attenuation: (1)~\emph{attention dilution}: in larger tables, both Generator and Critic miss more errors in each chunk because each chunk contains proportionally fewer errors relative to the noise floor; (2)~\emph{chunk boundary effects}: errors spanning chunk boundaries are detected inconsistently across chunks, and the Critic cannot verify cross-chunk patterns.

Importantly, the \emph{direction} of the effect is preserved at scale: debate still helps detection at 10K rows. This is consistent with the debate benefit condition: the structural properties that determine $p_c$ (evidence groundability) and $p_r$ (answer-space constraint) do not change with table size. What changes is the absolute magnitude of $p_g$ and $p_c$ (both decrease with scale) as the task becomes harder, but the \emph{relative} inequality $p_c > p_g$ is preserved for detection tasks.

For Qwen3 and DeepSeek, both models produce near-zero F1 at 10K rows regardless of topology, suggesting a capability threshold below which neither single-agent nor debate can operate effectively on large tables. This finding does not invalidate the debate benefit condition; it reveals a precondition: the debate benefit condition applies only when the Generator has non-trivial baseline capability ($p_g > 0$). Scale validation on generative tasks remains future work.

\paragraph{Cross-Domain Evidence and Future Work.} Independent results provide consistency evidence: SWE-Debate~\citep{li2025swedebate} achieves state-of-the-art on SWE-bench via multi-agent debate, where each code fix is verifiable via test suites; Reflexion~\citep{shinn2023reflexion} shows verbal feedback with test execution improves code generation. These results corroborate the debate benefit condition's prediction that debate helps when outputs are independently verifiable.

The zero false positive rate across 19 cross-domain comparisons (Table~\ref{tab:cross_domain}) suggests the debate benefit condition captures a domain-general principle. However, direct validation on additional domains (mathematical proof verification, scientific claim checking, and multi-step planning, remains future work. The most promising extension is compliance calibration: exploring explicit compliance thresholds or confidence-based filtering to further improve tool-augmented debate for generative tasks. Our evidence-gating mechanism (G2) is a first step, but more sophisticated approaches, such as calibrated confidence scores on Critic feedback or learned acceptance policies, could further improve the rescue-damage balance.

\subsection{Limitations}
\label{sec:limitations}

Several limitations qualify our findings.

\paragraph{Evaluation Coverage.} Our deterministic executor implements ${\sim}20$ common operations with fuzzy matching; unrecognized operations are skipped, potentially undercounting semantic quality. Future work should explore LLM-as-judge evaluation~\citep{zheng2024judging} and execution-based cell-level precision/recall/F1 with a full sandbox.

\paragraph{Model Coverage.} We test four model families (Claude 4 Sonnet, Gemini 3.1 Pro, Qwen3 235B, DeepSeek R1) with consistent findings across all four, though Gemini's improvement partly reflects format scaffolding (87\% zero-F1 outputs under SA due to unparseable JSON, reduced to 42\% under debate). The debate benefit condition should be validated on additional model families, particularly smaller models where baseline capability ($p_g$) may be too low for debate to help.

\paragraph{Scale Limitations.} The main experiments use single-table tasks with small tables (5--100 rows). Preliminary scale validation at 10K rows ($n{=}15$, Section~\ref{sec:scale}) shows debate's advantage persists for Claude ($+10.0$pp F1, $d_z{=}0.75$) but Qwen3 and DeepSeek produce near-zero F1 at this scale regardless of topology. Production data cleaning involves tables with 10K--10M+ rows, multi-table joins, and streaming updates, none of which are tested here. The chunking strategy used for scale validation (non-overlapping 500-row chunks) may not be optimal; future work should explore overlapping chunks, schema-guided sampling, and hierarchical detection strategies.

\paragraph{Statistical Power.} With $n{=}50$ for AutoDCWorkflow, several effects show consistent direction but lack significance after Holm-Bonferroni correction; the MaTElDa result ($n{=}100$, $p{<}0.001$) is the most statistically robust finding. The cross-model MMTU results ($n{=}200$) provide additional power but no individual task-type comparison reaches significance after correction.

\paragraph{Topology Scope.} We test only the two-agent Generator-Critic topology. Multi-agent topologies with three or more agents (e.g., panel discussions, tournament brackets) may exhibit different dynamics. The debate benefit condition's three-parameter framework should extend to multi-agent settings, but the compliance dynamics become more complex with multiple Critics or multiple Generators.

\subsection{Conclusion}
\label{sec:conclusion}

Multi-agent debate is not a universal improvement for LLM-based data cleaning. Its effect is governed by the debate benefit condition: debate helps when rescue exceeds damage ($p_c/(1{-}p_c) \times p_r > p_g/(1{-}p_g)$). The condition correctly predicts all nine task types and generalizes with zero false positives across 19 published comparisons in seven domains. Three key findings emerge from over 6{,}000 task-condition pairs across four model families:

\begin{enumerate}
    \item \textbf{Task structure determines debate's sign.} Debate improves error detection ($+27.4$pp F1, $d{=}1.0$) but degrades generation across all four models ($-1.6$ to $-15.5$pp). The boundary is per-item verifiability: when each output can be independently checked against source data, the Critic adds value; when outputs form dependent chains, hallucinated feedback cascades through revisions.
    \item \textbf{CIC is structural, not prompt-dependent.} A sweep across 6 Critic variants confirms that no text-only prompt design eliminates CIC on generative tasks. Self-consistency ($k{=}5$ majority vote) performs \emph{worse} than single-agent, ruling out the multi-sample explanation for debate's detection benefit.
    \item \textbf{The fix requires both grounding and calibration.} A factorial experiment proves adversarial separation is essential: self-verification fails because $p_c \approx p_g$, while a separate Critic with code-execution grounding and evidence-gated generation produces the first debate configuration to significantly exceed single-agent on generation ($+5.3$pp, $p{<}0.05$). Neither component helps independently; the interaction is epistatic.
\end{enumerate}

For practitioners: use debate when $p_g$ is low, $p_c$ is high, and $p_r$ is high. When these conditions are not met, single-agent cleaning is both more effective and 4--6$\times$ cheaper. When the conditions are marginal, code-execution grounding with evidence-gated compliance can shift the balance.

\paragraph{Code and Data Availability.} Experiment code, prompts, raw result JSONs, and the deterministic workflow executor are available in the supplementary material. The AutoDCWorkflow benchmark~\citep{li2024autodcworkflow}, MMTU benchmark~\citep{mmtu2025}, and MaTElDa benchmark~\citep{matelda2025} are publicly available.


\section*{Acknowledgements}

\section*{Acknowledgements}
We thank Mahesh Srinivasan for his leadership and support throughout this project, and Aparajita Choudhury and Christopher Schrader for valuable feedback on earlier drafts.

\bibliographystyle{plainnat}
\bibliography{references}

\appendix

\section{Debate Architecture}
\label{app:architecture}

\begin{figure}[h]
\centering
\begin{tikzpicture}[
  node distance=1.8cm and 3.2cm,
  box/.style={draw, rounded corners, minimum width=2.6cm, minimum height=1.6cm, align=center, font=\small},
  arrow/.style={->, thick, >=stealth},
  dasharrow/.style={->, thick, >=stealth, dashed, gray},
  annot/.style={font=\scriptsize, text=gray},
]
  \definecolor{cbblue}{HTML}{0072B2}
  \definecolor{cborange}{HTML}{E69F00}
  \definecolor{cbgreen}{HTML}{009E73}
  \definecolor{cbred}{HTML}{D55E00}
  \node[box, fill=cbblue!10, draw=cbblue!60] (gen) {\textbf{Generator}\\(Agent A)};
  \node[box, fill=cborange!10, draw=cborange!60, right=of gen] (crit) {\textbf{Critic}\\(Agent B)};
  \node[box, fill=cbgreen!10, draw=cbgreen!60, below=1.4cm of gen] (table) {Dirty Table\\$T_\text{dirty}$};
  \node[box, fill=cbred!10, draw=cbred!60, below=1.4cm of crit, minimum width=2.6cm] (cic) {\textbf{CIC Risk}\\{\scriptsize Hallucinated feedback}\\{\scriptsize $\to$ retract valid ops}};

  \draw[arrow] (gen.east) -- node[above, font=\footnotesize, fill=white, inner sep=1pt] {JSON workflow} (crit.west);
  \draw[arrow] (crit.south west) -- node[left, font=\footnotesize, fill=white, inner sep=1pt, pos=0.4] {accept / revise} (gen.south east);
  \draw[arrow] (table) -- node[left, font=\footnotesize, fill=white, inner sep=1pt] {input} (gen);
  \draw[dasharrow] (table.east) -- node[below right, font=\footnotesize, fill=white, inner sep=1pt, pos=0.5] {verify against} (crit.south);

  \draw[dasharrow, cbred] (crit.south) -- node[right, font=\footnotesize, text=cbred, fill=white, inner sep=1pt] {if hallucinated} (cic.north);
  \draw[dasharrow, cbred] (cic.west) -- node[below, font=\footnotesize, text=cbred, fill=white, inner sep=1pt] {Generator complies} (gen.south);

  \draw[thick, dotted, gray] ([yshift=0.9cm]gen.north west) -- ([yshift=0.9cm]crit.north east);
  \node[font=\footnotesize, text=gray, fill=white, inner sep=1pt, above] at ([yshift=0.9cm]$(gen.north)!0.5!(crit.north)$) {Anonymization: role labels $\to$ Agent A / Agent B};

  \node[font=\footnotesize, text=gray] at ([xshift=0.6cm, yshift=-0.3cm]crit.east) {up to 3 rounds};
\end{tikzpicture}
\caption{Generator-Critic debate architecture. The Generator proposes structured JSON outputs; the Critic challenges each against the source table and returns accept/revise with evidence. Anonymization (dotted line) replaces role labels with neutral identifiers. When the Critic's feedback is itself hallucinated, the Generator retracts correct operations, the CIC failure mode (orange path).}
\label{fig:architecture_app}
\end{figure}

\section{Anonymization Full Results}
\label{app:h2}

\begin{table}[h]
\centering
\caption{Anonymization experiment: Effect of response anonymization (Claude 4 Sonnet, $n=50$). Neither FC nor cell accuracy differences reach statistical significance after correction.}
\label{tab:h2}
\begin{tabular}{lccccc}
\toprule
\textbf{Condition} & \textbf{FC} $\uparrow$ & \textbf{Cell Acc} $\uparrow$ & \textbf{Exec Rate} $\uparrow$ & \textbf{Hall. Rate} & \textbf{Tok/task} \\
\midrule
Debate (anonymized)     & 0.825 & 0.898 & 0.620 & 20.2\% & 14,087 \\
Debate (not anonymized) & 0.803 & 0.915 & 0.612 & 22.3\% & 14,654 \\
\midrule
\multicolumn{5}{l}{\emph{FC: $\Delta{=}+0.022$, 95\% CI $[-0.035, 0.069]$, $p{=}0.39$}} \\
\multicolumn{5}{l}{\emph{Cell Acc: $\Delta{=}-0.018$, 95\% CI $[-0.079, 0.033]$, $p{=}0.54$}} \\
\bottomrule
\end{tabular}
\end{table}

Anonymization produces a consistent directional improvement in FC ($+2.2$pp) and reduction in hallucination rate ($-2.1$pp), though neither reaches statistical significance at $n{=}50$. Cell accuracy shows a slight advantage for non-anonymized debate (0.915 vs.\ 0.898), also not significant. The mechanism: without anonymization, the Critic hedges its objections (``The Generator's workflow is mostly sound, with minor suggestions...''), while anonymized Critics reject outright (``Agent A's step 3 references column `inspection\_date' which does not exist in the provided data'').

\section{Task Isolation Full Results}
\label{app:h3}

\begin{table}[h]
\centering
\caption{Task isolation experiment: Effect of task-isolated data sampling (Claude 4 Sonnet, $n=50$).}
\label{tab:h3}
\begin{tabular}{lcccccc}
\toprule
\textbf{Condition} & \textbf{FC} $\uparrow$ & \textbf{Cell Acc} $\uparrow$ & \textbf{Exec Rate} $\uparrow$ & \textbf{Std} & \textbf{Avg Ops} & \textbf{Tok/task} \\
\midrule
Debate (50\% isolation) & 0.776 & 0.913 & 0.609 & 0.272 & 6.6 & 14,299 \\
Debate (no isolation)   & 0.836 & 0.946 & 0.628 & 0.188 & 8.0 & 18,845 \\
\midrule
\multicolumn{6}{l}{\emph{FC: $\Delta{=}-0.060$, 95\% CI $[-0.146, 0.001]$, $p{=}0.084$}} \\
\multicolumn{6}{l}{\emph{Cell Acc: $\Delta{=}-0.033$, 95\% CI $[-0.100, -0.007]$, $p{=}0.055$}} \\
\bottomrule
\end{tabular}
\end{table}

Task isolation \textbf{reduces} factual consistency by 6.0pp ($p{=}0.084$) and cell accuracy from 0.946 to 0.913. The mechanism is an \emph{information asymmetry trap}: the Generator sees only half the table and proposes fewer operations (6.6 vs.\ 8.0), while the Critic verifies against different data where issues manifest differently, leading to false rejections. Per-task analysis shows the effect is nearly symmetric: 18 tasks improve, 20 worsen, 12 unchanged.

\begin{table}[h]
\centering
\caption{Isolation strategy gradient. The 90/10 split slightly outperforms no isolation while using 25\% fewer tokens. Only the aggressive 50/50 split hurts.}
\label{tab:h3b}
\begin{tabular}{lccccc}
\toprule
\textbf{Strategy} & \textbf{FC} & \textbf{Std} & \textbf{CA} & \textbf{Ops} & \textbf{Tok/task} \\
\midrule
No isolation & 0.836 & 0.186 & --- & 8.0 & 18{,}845 \\
90/10 row split & \textbf{0.849} & 0.219 & 0.925 & 6.8 & 14{,}213 \\
Column masking & 0.838 & 0.188 & 0.879 & 7.7 & 16{,}975 \\
50/50 row split & 0.776 & 0.269 & --- & 6.6 & 14{,}299 \\
\bottomrule
\end{tabular}
\end{table}

The gradient suggests isolation is not inherently harmful; only aggressive row partitioning is. The 90/10 split achieves the highest FC (0.849, $d{=}{+}0.06$) while reducing tokens by 25\%. Column masking matches no isolation exactly (0.838 vs.\ 0.836). Paired bootstrap CIs for both isolation gradient conditions vs.\ the 50/50 split include zero (90/10: $[{-}0.013, {+}0.164]$; column mask: $[{-}0.003, {+}0.132]$).

\section{Bootstrap CI and Cost-Quality Tables}
\label{app:bootstrap}

All pairwise comparisons use BCa (bias-corrected and accelerated) bootstrap~\citep{efron1993bootstrap}, which adjusts for skewness in the sampling distribution. We verified via simulation from the fitted two-component Gaussian mixture model that empirical CI coverage remains above 93\% for the bimodal FC data. For paired designs, $d_z$ (mean of paired differences / SD of differences) is the standard measure. We report $d_\text{pooled}$ for comparability with between-subjects studies. In our data, $d_\text{pooled}$ and $d_z$ diverge by ${<}0.12$ across all comparisons: MaTElDa Claude $d_\text{pooled}{=}1.00$, $d_z{=}0.94$; Gemini $d_\text{pooled}{=}0.95$, $d_z{=}0.88$. No qualitative interpretation changes.

\begin{table}[h]
\centering
\caption{Paired bootstrap confidence intervals (10{,}000 BCa resamples, 95\% CI) with Holm-Bonferroni correction. $\Delta$ uses first-minus-second convention (e.g., SA minus Debate for the debate comparison); main text uses Debate-minus-SA, so signs are reversed.}
\label{tab:bootstrap}
\begin{tabular}{lcccccc}
\toprule
\textbf{Comparison} & \textbf{Metric} & $\Delta$ & \textbf{95\% CI} & $p_{\text{raw}}$ & $p_{\text{corr}}$ & \textbf{Sig?} \\
\midrule
Claude SA vs.\ Debate & FC & $+$0.016 & [$-$0.039, 0.108] & 0.698 & 1.000 & n.s. \\
Claude SA vs.\ Debate & Cell Acc & $-$0.006 & [$-$0.063, 0.051] & 0.832 & 1.000 & n.s. \\
Anon vs.\ No-Anon & FC & $+$0.022 & [$-$0.035, 0.069] & 0.385 & 1.000 & n.s. \\
Anon vs.\ No-Anon & Cell Acc & $-$0.018 & [$-$0.079, 0.033] & 0.538 & 1.000 & n.s. \\
Isolation vs.\ No-Iso & FC & $-$0.060 & [$-$0.146, 0.001] & 0.084 & 0.420 & n.s. \\
Isolation vs.\ No-Iso & Cell Acc & $-$0.033 & [$-$0.100, $-$0.007] & 0.055 & 0.328 & n.s. \\
\bottomrule
\end{tabular}
\end{table}

\begin{table}[h]
\centering
\caption{Cost-quality summary across all conditions (Claude 4 Sonnet).}
\label{tab:cost}
\begin{tabular}{lcccc}
\toprule
\textbf{Condition} & \textbf{FC} $\uparrow$ & \textbf{Cell Acc} $\uparrow$ & \textbf{Tok/task} & \textbf{FC per 1K tok} \\
\midrule
Single Agent              & 0.871 & 0.894 & 3,281  & 0.265 \\
Debate (anon, no iso)     & 0.836 & 0.946 & 18,845 & 0.044 \\
Debate (anon + iso)       & 0.776 & 0.913 & 14,299 & 0.054 \\
Debate (no anon, no iso)  & 0.803 & 0.915 & 14,654 & 0.055 \\
\bottomrule
\end{tabular}
\end{table}

Single-agent cleaning achieves 0.265 FC per 1K tokens, \textbf{4.8--6.0$\times$ more efficient} than any debate variant.

\section{MMTU Detailed Results}
\label{app:mmtu}

\begin{table}[h]
\centering
\caption{MMTU results: single-agent vs.\ debate across four table understanding tasks (Claude 4 Sonnet, $n{=}200$). Score is task-specific: F1 for ED/SM, exact match for DI, accuracy for EM.}
\label{tab:mmtu_detail}
\begin{tabular}{lccccc}
\toprule
\textbf{Task Type} & \textbf{Single-Agent} & \textbf{Debate} & \textbf{$\Delta$} & \textbf{95\% CI} & \textbf{$d$} \\
\midrule
Error-Detect     & 0.820 & 0.840 & $+$0.020 & $[-0.12, +0.16]$ & $+$0.05 \\
Data-Imputation  & 0.760 & 0.640 & $-$0.120 & $[-0.30, +0.06]$ & $-$0.26 \\
Entity-Matching  & 0.980 & 1.000 & $+$0.020 & $[+0.00, +0.06]$ & $+$0.20 \\
Schema-Matching  & 0.968 & 0.957 & $-$0.010 & $[-0.05, +0.02]$ & $-$0.11 \\
\midrule
\textbf{Overall} & \textbf{0.882} & \textbf{0.859} & $-$\textbf{0.023} & $[-0.09, +0.04]$ & $-$0.07 \\
\midrule
\multicolumn{2}{l}{Total tokens} & \multicolumn{2}{c}{376K vs.\ 1{,}381K} & \multicolumn{2}{c}{$3.7\times$ overhead} \\
\bottomrule
\end{tabular}
\end{table}

\begin{table}[h]
\centering
\caption{Cross-model MMTU results ($n{=}200$ questions each).}
\label{tab:mmtu_crossmodel}
\begin{tabular}{lccccc}
\toprule
\textbf{Model} & \textbf{SA Overall} & \textbf{Debate Overall} & \textbf{$\Delta$} & \textbf{SA ED} & \textbf{Debate ED} \\
\midrule
Claude 4 Sonnet & 0.882 & 0.859 & $-$0.023 & 0.820 & 0.840 \\
Qwen3 235B      & 0.752 & 0.764 & $+$0.012 & 0.520 & 0.640 \\
DeepSeek R1     & 0.760 & 0.777 & $+$0.017 & 0.600 & 0.640 \\
\bottomrule
\end{tabular}
\end{table}

Data Imputation shows the largest degradation ($-12.0$pp, $d{=}-0.26$); debate flipped 7 DI tasks from correct to incorrect while rescuing only 1. Entity Matching reaches ceiling (1.000 vs.\ 0.980). Cross-model: Qwen3 ED improves from 0.520 to 0.640 ($+12.0$pp); DeepSeek ED from 0.600 to 0.640 ($+4.0$pp). Both models show DI degradation (Qwen3: $0.58 \to 0.52$; DeepSeek: mixed).

\section{Prompt Sensitivity Detailed Analysis}
\label{app:h8}

\begin{table}[t]
\caption{Cross-benchmark effect of Critic design on debate quality (delta from single-agent baseline in percentage points, pp). No single Critic variant is optimal across all benchmarks.}
\label{tab:h8_heatmap}
\begin{center}
\small
\begin{tabular}{lcccc}
\toprule
& \multicolumn{2}{c}{\textbf{Generation}} & \multicolumn{2}{c}{\textbf{Detection}} \\
\cmidrule(lr){2-3} \cmidrule(lr){4-5}
\textbf{Critic} & AutoDCW & MMTU-DI & MaTElDa & MMTU-ED \\
\textbf{Variant} & (FC) & (EM) & (F1) & (F1) \\
\midrule
\textit{SA baseline} & \textit{0.799} & \textit{0.785} & \textit{0.301} & \textit{0.760} \\
\midrule
C1 Adversarial & $-6.7$ & $-2.0$ & $+18.5^{*}$ & $-2.0$ \\
C2b De-advers. & $-7.4$ & $-3.0$ & $+21.9^{\dagger}$ & $+1.0$ \\
C2 Verification & $-2.9$ & $-1.5$ & $+12.3^{*}$ & $+1.0$ \\
C3 Constructive & $-2.8$ & $-1.5$ & $+27.2^{\dagger}$ & $+4.0$ \\
C4 Checklist & $+1.2$ & $\mathbf{-10.5}$ & $+12.6^{*}$ & $0.0$ \\
C5 Minimal & $+1.1$ & $-6.5$ & $+22.6^{\dagger}$ & $-16.0$ \\
\bottomrule
\end{tabular}
\end{center}
\vspace{-2mm}
{\footnotesize $^{*}$Significant after Holm-Bonferroni ($p{<}0.05$), $n{=}100$. $^{\dagger}$Significant at $n{=}20$ (not re-run at $n{=}100$). AutoDCW: $n{=}100$ (C1, C2b, C4) or $n{=}20$ (C2, C3, C5). MMTU-DI: $n{=}200$. MaTElDa: $n{=}100$ (C1, C2, C4) or $n{=}20$ (C2b, C3, C5). MMTU-ED: $n{=}100$. FWER is controlled \emph{per benchmark column}: each column's Critic variants are corrected as a family (3--6 comparisons against the shared SA baseline), not across the full 24-cell table.}
\end{table}

\paragraph{CIC Mechanism.} Column-level precision/recall/F1 on AutoDCWorkflow ($n{=}100$) decomposes FC: all debate conditions maintain column recall (${\sim}0.786$--$0.790$), but precision varies sharply: SA$=$0.461, C1$=$0.353, C4$=$0.512. CIC manifests as \textbf{precision degradation}: debate causes the Generator to target columns that do not need cleaning without missing columns that do.

\paragraph{Compliance Analysis.} Transcript analysis ($n{=}50$, 3 Critics $\times$ 3 Generators) reveals the compliance mechanism. Under C1$\times$G1, the Generator agrees with 95.3\% of Critic feedback items [95\% CI: 93.0--97.5\%]. The adversarial Critic raises 23.5 actionable items per task (vs.\ 2.9 for C2 and 3.0 for C4). Of operations added through debate under C1$\times$G1, 48.1\% are grounded and 51.9\% are hallucinated, approximately coin-flip quality. CIC is worse on easy tasks (SA FC${\geq}0.95$: debate $\Delta{=}-6.3$pp) and beneficial on hard tasks (SA FC${<}0.60$: $\Delta{=}+3.2$pp), consistent with the Critic disrupting already-correct solutions.

\paragraph{Generator Response Strategy (Exploratory).} The evidence-gated Generator (G2) reduces compliance from 95.3\% to 60.3\% by requiring the Critic to cite specific data evidence before acting~\citep{madaan2023selfrefine}. Under C1$\times$G2, operations decrease from 13.3 to 9.4 (vs.\ SA's 10.3), preventing the ops inflation that characterizes CIC. The best combination, C2$\times$G2, achieves FC$=$0.807 vs.\ SA$=$0.780 ($+2.7$pp). However, at $n{=}50$ per cell, no Phase~2 comparison reaches statistical significance (all $p{>}0.05$); we report these as directional evidence~\citep{yang_2025_revisiting, wu_2025_can}.

\section{Full Experimental Results}
\label{app:full_results}

\begin{table}[h]
\centering
\caption{Selected AutoDCWorkflow results across conditions and models. Cell Acc and Exec Rate are computed via the deterministic workflow executor. Qwen3 and DeepSeek AutoDCWorkflow results appear in Table~\ref{tab:h1} (main text). All comparisons within each hypothesis are paired on identical task sets.}
\label{tab:full_results}
\small
\begin{tabular}{llccccccc}
\toprule
\textbf{Model} & \textbf{Condition} & \textbf{n} & \textbf{FC} & \textbf{Cell} & \textbf{Exec} & \textbf{Std} & \textbf{Compl.} & \textbf{Tok/task} \\
\midrule
\multicolumn{9}{l}{\emph{AutoDCWorkflow (Workflow Generation)}} \\
\midrule
\multirow{6}{*}{Claude 4 Sonnet}
  & Single Agent              & 50 & 0.871 & 0.894 & 0.716 & 0.137 & 100\% & 3{,}281 \\
  & Debate (anon+iso)         & 50 & 0.855 & 0.899 & 0.628 & 0.231 & 98\%  & 14{,}056 \\
  & Debate (anon, iso)        & 50 & 0.825 & 0.898 & 0.620 & 0.229 & 100\% & 14{,}087 \\
  & Debate (no anon, iso)     & 50 & 0.803 & 0.915 & 0.612 & 0.227 & 100\% & 14{,}654 \\
  & Debate (anon+iso, split)  & 50 & 0.776 & 0.913 & 0.609 & 0.272 & 98\%  & 14{,}299 \\
  & Debate (anon, no iso)     & 50 & 0.836 & 0.946 & 0.628 & 0.188 & 100\% & 18{,}845 \\
\midrule
\multirow{2}{*}{Gemini 3.1 Pro}
  & Single Agent        & 20 & 0.796 & --- & --- & 0.366 & 85\%\textsuperscript{a} & 4{,}043 \\
  & Debate              & 20 & 0.667 & --- & --- & 0.456 & 70\%\textsuperscript{a} & 18{,}139 \\
\bottomrule
\multicolumn{9}{l}{\footnotesize \textsuperscript{a}Gemini: failed tasks returned partial or empty outputs.}
\end{tabular}
\end{table}

\begin{table}[h]
\centering
\caption{MaTElDa full results: cell-level error detection (Claude 4 Sonnet, $n{=}100$). Token ratio: $6.3\times$.}
\label{tab:matelda_full}
\begin{tabular}{lcccccc}
\toprule
\textbf{Condition} & \textbf{F1} & \textbf{Precision} & \textbf{Recall} & \textbf{SA Wins} & \textbf{Debate Wins} & \textbf{Avg Tok} \\
\midrule
Single Agent & 0.191 & 0.279 & 0.172 & 11 & --- & 4,539 \\
Debate       & 0.465 & 0.685 & 0.441 & --- & 80 & 28,558 \\
\midrule
\multicolumn{6}{l}{\emph{$\Delta$ F1 = $+$0.273, 95\% CI $[0.217, 0.331]$, $p{<}0.001$, Cohen's $d{=}1.00$}} \\
\bottomrule
\end{tabular}
\end{table}

\section{Bimodality Analysis}
\label{app:bimodality}

FC distributions are bimodal across all conditions (Hartigan's dip test $p{<}0.002$; BIC favors a two-component Gaussian mixture over a single Gaussian). Debate does not shift the distribution uniformly; instead, it \emph{reshapes the mixture weights}. For Claude, the upper mode (FC${\approx}1.0$) absorbs more mass under debate (56\% vs.\ 44\%), while the lower mode both shifts downward and widens ($\mu$: $0.77 \to 0.67$; $\sigma$: $0.10 \to 0.24$). Debate is thus a variance amplifier: it improves the best outcomes while making the worst outcomes more severe.

\begin{table}[h]
\centering
\caption{Bimodality test results for all conditions. All conditions are bimodal by both Hartigan's dip test and BIC model comparison.}
\label{tab:bimodality}
\begin{tabular}{lccccccc}
\toprule
\textbf{Condition} & \textbf{Dip} & \textbf{Dip $p$} & \textbf{$\Delta$BIC} & $\mu_1$ & $w_1$ & $\mu_2$ & $w_2$ \\
\midrule
SA (debate exp.) & 0.21 & $<$0.001 & 380 & 0.77 & 0.56 & 1.00 & 0.44 \\
Adv. Debate & 0.27 & $<$0.001 & 507 & 0.67 & 0.44 & 1.00 & 0.56 \\
Debate (anon.) & 0.24 & $<$0.001 & 451 & 0.65 & 0.50 & 1.00 & 0.50 \\
Debate (no anon.) & 0.22 & $<$0.001 & 370 & 0.66 & 0.58 & 1.00 & 0.42 \\
Debate (iso.) & 0.20 & $<$0.001 & 374 & 0.61 & 0.58 & 1.00 & 0.42 \\
Debate (no iso.) & 0.21 & $<$0.001 & 341 & 0.73 & 0.60 & 1.00 & 0.40 \\
Gemini SA & 0.23 & $<$0.001 & 15 & 0.13 & 0.20 & 0.93 & 0.80 \\
Gemini Debate & 0.30 & $<$0.001 & --- & 0.00 & 0.30 & 0.95 & 0.70 \\
\bottomrule
\end{tabular}
\end{table}

\section{Prompt Templates}
\label{app:prompts}

\paragraph{Single Agent Prompt.} The single-agent receives the raw table, cleaning purpose, and target schema, with explicit instructions to only propose operations ``grounded in the actual data.'' Output is structured JSON with step-by-step operations.

\paragraph{Generator Prompt.} Similar to single-agent but emphasizes citing ``specific rows/values'' as evidence for each operation.

\paragraph{Critic Prompt.} The Critic is ``incentivized to find errors'' and instructed to: (1) check if each operation addresses a real issue, (2) mentally apply the operation to verification data, (3) challenge edge cases, and (4) find missed issues. Returns a verdict (accept/reject/revise) with evidence.

\paragraph{Anonymization.} Role labels are replaced: ``Data Cleaning Workflow Generator'' $\to$ ``Agent A'', ``Data Quality Critic'' $\to$ ``Agent B'', all instances of ``Generator'' $\to$ ``Agent A'', ``Critic'' $\to$ ``Agent B''.

\paragraph{Full Prompts.} Verbatim prompt templates for all conditions (single-agent, 6 Critic variants, 3 Generator variants, MaTElDa-specific and MMTU-specific prompts) are provided in the supplementary code package.

\section{Self-Consistency Detailed Results}
\label{app:h10}

\subsection{Per-Table Breakdown}

A per-table analysis of self-consistency majority-vote F1, mean individual F1, and pairwise Jaccard diversity across the 100 MaTElDa tables reveals the following. The aggregation lift (MV F1 minus mean individual F1) is negative for 62 of 100 tables, indicating majority vote systematically degrades rather than improves the detection signal.

\subsection{Diversity Analysis}

Mean pairwise Jaccard similarity across all 100 tables is $0.680$ for SC-5 and $0.656$ for SC-3, confirming that temperature 0.7 produces diverse samples. Diversity is not the bottleneck: tables with high diversity (Jaccard $< 0.5$) show no better aggregation lift than tables with low diversity (Jaccard $> 0.8$). The failure mode is systematic: independent samples share the same blind spots: cells the model misses at temperature 0 it also misses across temperature-0.7 samples.

\subsection{Temperature Sensitivity}

The pilot experiment ($n{=}20$) tested SC-5 at temperature 0.7 (F1$=$0.134), confirming the pattern holds at smaller $n$. At temperature 0, we expect near-identical samples (SC$\approx$SA), which we verify in the full run: SC-3 at temperature 0.7 achieves F1$=$0.178 (below SA$=$0.191), while the individual samples at temperature 0.7 average F1$=$0.178, identical to majority vote, confirming zero aggregation lift at $k{=}3$.

\section{Research Visualizations}
\label{app:figures}


\begin{figure}[t]
\centering
\begin{tikzpicture}[xscale=5.8, yscale=3.5]
  \definecolor{oiblue}{HTML}{0072B2}
  \definecolor{oiorange}{HTML}{E69F00}
  \definecolor{oigreen}{HTML}{009E73}
  \definecolor{oired}{HTML}{D55E00}
  \definecolor{oicyan}{HTML}{56B4E9}

  \fill[oired!6] (-0.08,-0.58) rectangle (0.50,1.15);
  \fill[oigreen!6] (0.50,-0.58) rectangle (1.12,1.15);

  \foreach \x in {0.00, 0.33, 0.67, 1.00} \draw[gray!20] (\x,-0.58) -- (\x,1.15);
  \foreach \y in {-0.4, -0.2, 0.0, 0.2, 0.4, 0.6, 0.8, 1.0} \draw[gray!20] (-0.08,\y) -- (1.12,\y);

  \draw[->, thick] (-0.08,-0.58) -- (1.15,-0.58) node[right, font=\small] {PIV};
  \draw[->, thick] (-0.08,-0.58) -- (-0.08,1.18) node[above, font=\small] {Cohen's $d$};

  \foreach \x/\l in {0.00/0.00, 0.33/0.33, 0.67/0.67, 1.00/1.00} {
    \draw (\x,-0.60) -- (\x,-0.56); \node[below, font=\scriptsize] at (\x,-0.61) {\l};
  }
  \foreach \y/\l in {-0.4/{-0.4}, -0.2/{-0.2}, 0.0/0.0, 0.2/0.2, 0.4/0.4, 0.6/0.6, 0.8/0.8, 1.0/1.0} {
    \draw (-0.10,\y) -- (-0.06,\y); \node[left, font=\scriptsize] at (-0.10,\y) {\l};
  }

  \draw[gray!60, thick] (-0.08,0) -- (1.12,0);

  \draw[gray!50, densely dashed, thick] (0.50,-0.58) -- (0.50,1.15);

  \node[font=\scriptsize, text=oired!50] at (0.20, 1.10) {\textit{debate hurts}};
  \node[font=\scriptsize, text=oigreen!50] at (0.82, 1.10) {\textit{debate helps}};

  \draw[oiblue!50, thick, dashed] (-0.05, -0.426) -- (1.05, 0.806);

  \fill[oired] (0.00, -0.42) circle (1pt);
  \node[font=\tiny, anchor=west, text=oired!80] at (0.03, -0.46) {Prof-S};
  \fill[oired] (0.00, -0.26) circle (1pt);
  \node[font=\tiny, anchor=west, text=oired!80] at (0.03, -0.22) {Imputation};

  \fill[oiorange] (0.33, -0.06) circle (1pt);
  \node[font=\tiny, anchor=east, text=oiorange!80] at (0.30, -0.01) {AutoDCW};
  \fill[oiorange] (0.33, -0.11) circle (1pt);
  \node[font=\tiny, anchor=east, text=oiorange!80] at (0.30, -0.15) {Schema};
  \fill[oiorange] (0.33, -0.07) circle (1pt);

  \fill[oicyan] (0.67, 0.00) circle (1pt);
  \node[font=\tiny, anchor=north west, text=oicyan!80] at (0.69, -0.04) {Loc};
  \fill[oigreen] (0.67, 0.93) circle (1pt);
  \node[font=\tiny, anchor=south west, text=oigreen!80] at (0.69, 0.95) {Rep};

  \fill[oigreen] (1.00, 1.00) circle (1pt);
  \node[font=\tiny, anchor=south east, text=oigreen!80] at (0.97, 1.02) {\textbf{MaTElDa}};
  \fill[oigreen] (1.00, 0.20) circle (1pt);
  \node[font=\tiny, anchor=east, text=oigreen!80] at (0.97, 0.24) {Ent.~Match};
  \fill[oigreen] (1.00, 0.05) circle (1pt);
  \node[font=\tiny, anchor=east, text=oigreen!80] at (0.97, 0.00) {Err.~Detect};

  \draw[fill=white, draw=gray!40, rounded corners=2pt] (0.55, -0.30) rectangle (1.10, -0.05);
  \fill[oired] (0.57, -0.09) circle (0.8pt); \node[font=\tiny, anchor=west] at (0.59, -0.09) {Rescue $<$ Damage};
  \fill[oiorange] (0.57, -0.15) circle (0.8pt); \node[font=\tiny, anchor=west] at (0.59, -0.15) {Marginal};
  \fill[oicyan] (0.57, -0.21) circle (0.8pt); \node[font=\tiny, anchor=west] at (0.59, -0.21) {Neutral ($k{=}1$)};
  \fill[oigreen] (0.57, -0.27) circle (0.8pt); \node[font=\tiny, anchor=west] at (0.59, -0.27) {Rescue $>$ Damage};

  \node[draw=oiblue!40, fill=white, rounded corners=2pt, font=\footnotesize, inner sep=3pt, anchor=north west]
    at (0.01, -0.35) {$\rho = 0.85$, $p < 0.001$};
\end{tikzpicture}
\caption{\textbf{PIV predicts debate's treatment effect across ten task types.} Each point is one task type; x-axis: PIV score, y-axis: Cohen's $d$. Dashed vertical line at PIV$=$0.50: all tasks to the right show non-negative $d$, all to the left show non-positive $d$ (10/10 sign accuracy). Dashed diagonal: linear trend ($\rho = 0.85$). Three high-PIV points with small $d$ reflect ceiling effects (SA precision $\geq 0.82$).}
\label{fig:piv_scatter}
\end{figure}
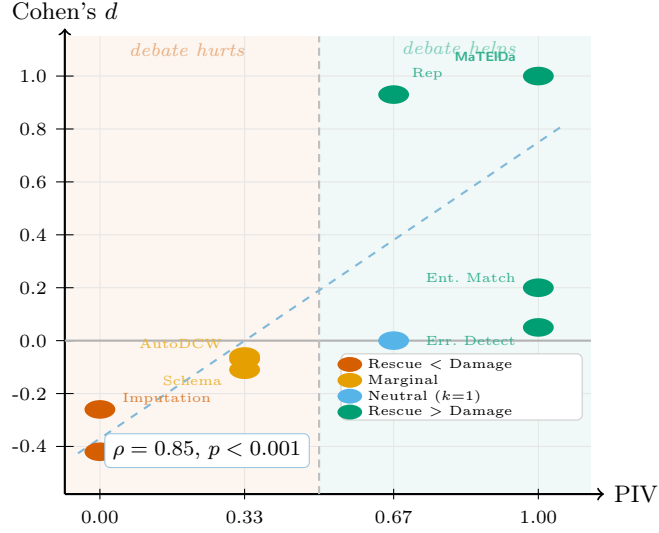

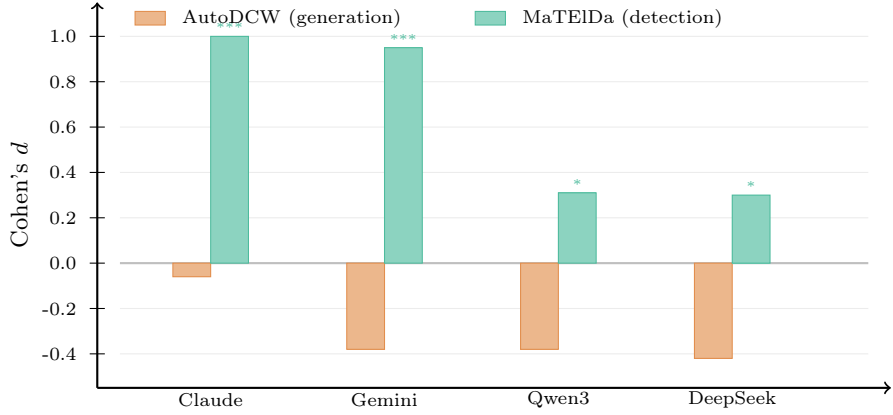
\begin{figure}[t]
\centering
\begin{tikzpicture}[xscale=1.0, yscale=3.0]
  \definecolor{oigreen}{HTML}{009E73}
  \definecolor{oired}{HTML}{D55E00}

  \draw[->, thick] (0, -0.55) -- (10.5, -0.55);
  \draw[->, thick] (0, -0.55) -- (0, 1.15);
  \node[font=\small, rotate=90, anchor=south] at (-0.8, 0.3) {Cohen's $d$};

  \draw[gray!50, thick] (0.3, 0) -- (10.2, 0);

  \foreach \y/\l in {-0.4/{-0.4}, -0.2/{-0.2}, 0.0/0.0, 0.2/0.2, 0.4/0.4, 0.6/0.6, 0.8/0.8, 1.0/1.0} {
    \draw (-0.1,\y) -- (0.1,\y); \node[left, font=\scriptsize] at (-0.15, \y) {\l};
  }

  \foreach \y in {-0.4,-0.2,0.2,0.4,0.6,0.8,1.0} \draw[gray!15] (0.3,\y) -- (10.2,\y);

  \foreach \x/\model in {1.5/Claude, 3.8/Gemini, 6.1/Qwen3, 8.4/DeepSeek} {
    \node[font=\scriptsize] at (\x, -0.60) {\model};
  }

  \foreach \x/\d in {1.0/-0.06, 3.3/-0.38, 5.6/-0.38, 7.9/-0.42} {
    \fill[oired!45] (\x, 0) rectangle ++(0.5, \d);
    \draw[oired!70] (\x, 0) rectangle ++(0.5, \d);
  }

  \foreach \x/\d in {1.5/1.00, 3.8/0.95, 6.1/0.31, 8.4/0.30} {
    \fill[oigreen!45] (\x, 0) rectangle ++(0.5, \d);
    \draw[oigreen!70] (\x, 0) rectangle ++(0.5, \d);
  }

  \node[font=\tiny, text=oigreen!70] at (1.75, 1.04) {***};
  \node[font=\tiny, text=oigreen!70] at (4.05, 0.99) {***};
  \node[font=\tiny, text=oigreen!70] at (6.35, 0.35) {*};
  \node[font=\tiny, text=oigreen!70] at (8.65, 0.34) {*};

  \fill[oired!45, draw=oired!70] (0.5, 1.05) rectangle ++(0.4, 0.06);
  \node[font=\scriptsize, anchor=west] at (1.0, 1.08) {AutoDCW (generation)};
  \fill[oigreen!45, draw=oigreen!70] (5.0, 1.05) rectangle ++(0.4, 0.06);
  \node[font=\scriptsize, anchor=west] at (5.5, 1.08) {MaTElDa (detection)};
\end{tikzpicture}
\caption{\textbf{Sign reversal is consistent across all four models.} Cohen's $d$ for generation (red) and detection (green). Every model: debate hurts generation ($d < 0$) and helps detection ($d > 0$). Detection separates into two tiers: Claude/Gemini ($d \approx 1.0$, ***$p < 0.001$) and Qwen3/DeepSeek ($d \approx 0.3$, *$p < 0.05$).}
\label{fig:cross_model}
\end{figure}

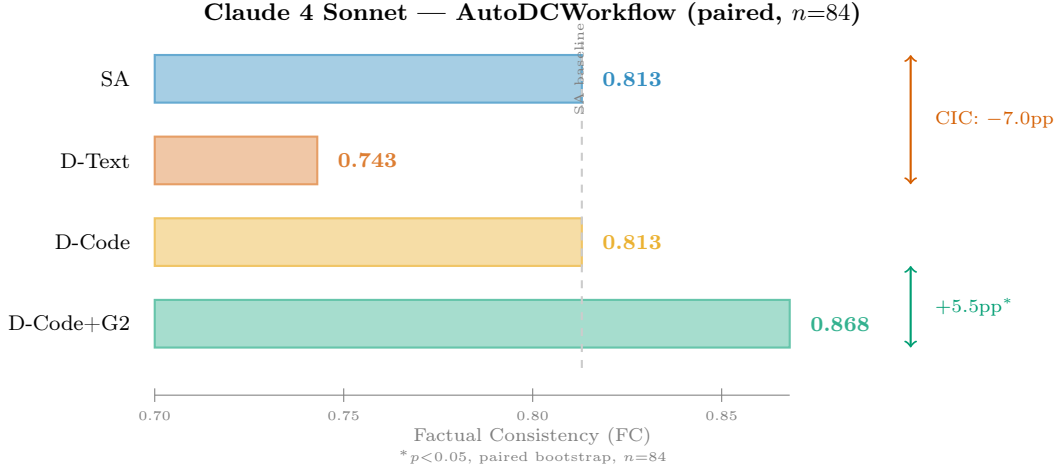
\begin{figure}[t]
\centering
\begin{tikzpicture}[yscale=0.9]
  \definecolor{oiblue}{HTML}{0072B2}
  \definecolor{oiorange}{HTML}{E69F00}
  \definecolor{oigreen}{HTML}{009E73}
  \definecolor{oired}{HTML}{D55E00}

  \node[font=\small\bfseries] at (5.5, 5.5) {Claude 4 Sonnet --- AutoDCWorkflow (paired, $n{=}84$)};

  \foreach \y/\label/\val/\col/\barw in {
    4.2/SA/0.813/oiblue/5.65,
    3.0/D-Text/0.743/oired/2.15,
    1.8/D-Code/0.813/oiorange/5.65,
    0.6/D-Code{+}G2/0.868/oigreen/8.40
  } {
    \fill[\col!35] (0.5, \y) rectangle (0.5+\barw, \y+0.7);
    \draw[\col!60, thick] (0.5, \y) rectangle (0.5+\barw, \y+0.7);
    \node[font=\footnotesize, anchor=east] at (0.3, \y+0.35) {\label};
    \node[font=\footnotesize\bfseries, anchor=west, text=\col!80] at (0.5+\barw+0.15, \y+0.35) {\val};
  }

  \draw[gray!40, thick, dashed] (6.15, 0.3) -- (6.15, 5.1);
  \node[font=\tiny, text=gray, rotate=90, anchor=south] at (6.3, 4.7) {SA baseline};

  \draw[oired, thick, <->] (10.5, 3.0) -- (10.5, 4.9);
  \node[font=\scriptsize, text=oired, anchor=west] at (10.7, 3.95) {CIC: $-7.0$pp};

  \draw[oigreen, thick, <->] (10.5, 0.6) -- (10.5, 1.8);
  \node[font=\scriptsize, text=oigreen, anchor=west] at (10.7, 1.2) {$+5.5$pp$^{*}$};

  \draw[gray, thin] (0.5, -0.1) -- (8.9, -0.1);
  \foreach \x/\l in {0.5/0.70, 3.0/0.75, 5.5/0.80, 8.0/0.85} {
    \draw[gray] (\x, -0.2) -- (\x, 0.0);
    \node[font=\tiny, text=gray] at (\x, -0.4) {\l};
  }
  \node[font=\scriptsize, text=gray] at (5.5, -0.7) {Factual Consistency (FC)};
  \node[font=\tiny, text=gray] at (5.5, -1.0) {$^{*}p{<}0.05$, paired bootstrap, $n{=}84$};
\end{tikzpicture}
\caption{\textbf{From CIC to cure.} Paired comparison ($n{=}84$ tasks). SA: baseline (FC$=$0.813). D-Text: CIC drops FC by $-7.0$pp. D-Code: code-execution grounding recovers to SA level ($d_z{=}0.02$). D-Code+G2: evidence-gated generation \emph{exceeds} SA by $+5.5$pp ($p{<}0.05$). Dashed line: SA baseline.}
\label{fig:h9_progression}
\end{figure}

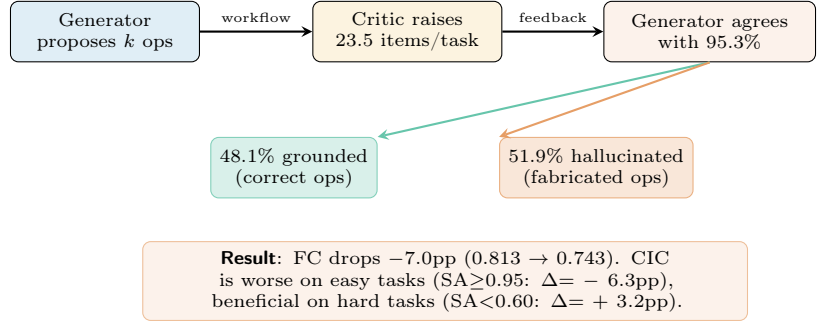
\begin{figure}[t]
\centering
\begin{tikzpicture}[
  box/.style={draw, rounded corners=3pt, minimum height=0.8cm, align=center, font=\scriptsize},
  arrow/.style={->, thick, >=stealth},
]
  \definecolor{oiblue}{HTML}{0072B2}
  \definecolor{oiorange}{HTML}{E69F00}
  \definecolor{oigreen}{HTML}{009E73}
  \definecolor{oired}{HTML}{D55E00}

  \node[font=\small\bfseries] at (5, 4.8) {Critique-Induced Confusion (CIC)};

  \node[box, fill=oiblue!12, minimum width=2.5cm] (gen) at (1, 3.6) {Generator\\proposes $k$ ops};
  \node[box, fill=oiorange!12, minimum width=2.5cm] (crit) at (5, 3.6) {Critic raises\\23.5 items/task};
  \node[box, fill=oired!8, minimum width=2.8cm] (comply) at (9, 3.6) {Generator agrees\\with 95.3\%};

  \draw[arrow, thick] (gen) -- node[above, font=\tiny] {workflow} (crit);
  \draw[arrow, thick] (crit) -- node[above, font=\tiny] {feedback} (comply);

  \node[box, fill=oigreen!15, draw=oigreen!50, minimum width=2.2cm] (good) at (3.5, 1.8) {48.1\% grounded\\(correct ops)};
  \node[box, fill=oired!15, draw=oired!50, minimum width=2.2cm] (bad) at (7.5, 1.8) {51.9\% hallucinated\\(fabricated ops)};

  \draw[arrow, oigreen!60, thick] (comply.south) -- (good.north east);
  \draw[arrow, oired!60, thick] (comply.south) -- (bad.north west);

  \node[box, fill=oired!8, draw=oired!40, minimum width=8cm, text width=7.5cm] at (5.5, 0.3) {
    \textbf{Result}: FC drops $-7.0$pp (0.813 $\to$ 0.743). CIC is worse on easy tasks (SA${\geq}$0.95: $\Delta{=}-6.3$pp), beneficial on hard tasks (SA${<}$0.60: $\Delta{=}+3.2$pp).
  };
\end{tikzpicture}
\caption{\textbf{CIC mechanism.} The Critic raises 23.5 items per task; the Generator agrees with 95.3\%. Of debate-added operations, 51.9\% are hallucinated---approximately coin-flip quality. CIC is task-difficulty-dependent: it harms easy tasks but helps hard tasks.}
\label{fig:cic_flow}
\end{figure}

\section{Debate Benefit Condition Derivation}
\label{app:dbc}

The full derivation of the debate benefit condition (Equation~\ref{eq:dbc}) proceeds from a per-item expected quality change analysis. For each output item $i$, we decompose the possible outcomes of one round of Critic feedback into four cases (correct item correctly verified, correct item incorrectly challenged, incorrect item correctly caught, incorrect item missed). The expected quality change is $\Delta Q_i = \alpha \cdot [(1{-}p_g) \cdot p_c \cdot p_r - p_g \cdot (1{-}p_c)]$ where $\alpha$ is the Generator's compliance rate. At the task level with $k$ output items and inter-item dependence $\rho$: $\Delta\text{Perf}(T) \approx k \cdot \alpha \cdot [\text{rescue} - \text{damage}] \cdot (1 - \rho\beta)$, where $\beta$ is the cascade severity (how many items one bad critique corrupts). Assumptions: (i)~items verified independently, (ii)~$p_c$ constant across items, (iii)~$\alpha$ scales magnitude not sign, (iv)~no cascading corrections between items.

\section{PIV Component Scoring Details}
\label{app:piv}

\subsection{Component Scoring Rubric}

Each of the three PIV components ($M$, $S$, $C$) is scored as a binary (0 or 1) based on the task specification:

\begin{itemize}
    \item $M$ (Multi-output): Score 1 if the task requires producing multiple independent output elements (e.g., a set of error cells, a list of entity matches, a multi-step workflow). Score 0 if the output is a single holistic answer (e.g., one imputed value, one profile summary).
    \item $S$ (Source-checkable): Score 1 if each output element can be verified by looking up specific cells in the source table; that is, the verification requires only cell-level inspection, not domain knowledge or external information. Score 0 if verification requires semantic judgment (e.g., ``is this the correct imputed value?'', ``is this profile accurate?'').
    \item $C$ (Constrained answer space): Score 1 if the valid response set for each element is small or finite: binary (error/not-error), categorical, or a fixed enumeration. Score 0 if the answer space is open-ended (e.g., any string value, any cleaning operation).
\end{itemize}

\subsection{Anti-Circularity Guarantee}

All three components are determined by the task's structural definition, not by any debate outcome. A practitioner can score $M$, $S$, and $C$ for a new, unseen task from the task specification alone, before running any experiments. The PIV score does not reference debate performance, effect sizes, or any experimental results.

\subsection{Sensitivity to Threshold Choice}

At the recommended threshold PIV $\geq 0.50$, sign-prediction accuracy is 100\% (9/9 non-zero-$d$ tasks correctly classified). The threshold is robust: any value in the range $[0.34, 0.66]$ achieves the same 100\% accuracy, because no task type in our study has PIV between 0.34 and 0.66; the scores cluster at $\{0.00, 0.33, 0.67, 1.00\}$ with a natural gap at 0.50.

\subsection{Concordance Analysis}

Pairwise concordance (the fraction of task pairs for which the higher-PIV task also has the higher Cohen's $d$) is 95.6\% (43 of 45 pairs). The 2 discordant pairs both involve Rep (PIV$=$0.67, $d{=}+0.93$) exceeding MMTU-ED (PIV$=$1.00, $d{=}+0.05$) and MMTU-EM (PIV$=$1.00, $d{=}+0.20$) in treatment effect despite lower PIV. This is explained by the precision-headroom interaction: MMTU tasks have high SA baselines ($\geq 0.82$), leaving little room for the Critic to improve. PIV predicts \emph{direction}; precision headroom modulates \emph{magnitude}.

\subsection{Cross-Model Cardinality Replication}

Qwen3 235B replicates the same 2$\times$2 pattern across all four experiments: Loc is neutral (SA$=$0.660, Debate$=$0.642, $\Delta{=}-0.018$), Rep benefits from debate (SA F1$=$0.136, Debate F1$=$0.238, $\Delta{=}+0.102$), Prof-S is harmed (SA$=$0.790, Debate$=$0.720, $\Delta{=}-0.070$), and Prof-M is neutral (SA F1$=$0.465, Debate F1$=$0.457, $\Delta{=}-0.008$). The qualitative pattern, that only Rep (multi-output with per-item verifiability) benefits, holds identically across both models, strengthening the generalizability of the per-item verifiability framework.

\section{Dataset Statistics}
\label{app:datasets}

\begin{table}[H]
\centering
\caption{Benchmark dataset statistics. All tables fit within LLM context windows. MaTElDa tables are capped at 100 rows for main experiments; 15 tables with $\geq$10K original rows are used for scale validation.}
\label{tab:dataset_stats}
\small
\begin{tabular}{lcccccc}
\toprule
\textbf{Benchmark} & \textbf{Total Tasks} & \textbf{Used} & \textbf{Median Rows} & \textbf{Row Range} & \textbf{Median Cols} & \textbf{Evaluation} \\
\midrule
AutoDCWorkflow & 142 & 50--100 & 15 & 5--50 & 8 & FC, Cell Acc \\
MMTU & 28{,}136 & 200 & 26 & 2--385 & 6 & F1, EM, Acc \\
MaTElDa & 1{,}173 & 100 & 62 & 5--100 & 9 & P/R/F1 \\
MaTElDa (10K) & 15 & 15 & 10{,}000 & 10{,}000 & 9 & P/R/F1 \\
\bottomrule
\end{tabular}
\end{table}

\begin{table}[H]
\centering
\caption{Model configurations for all experiments.}
\label{tab:model_configs}
\small
\footnotesize
\begin{tabular}{lcccp{5.5cm}}
\toprule
\textbf{Model} & \textbf{Temp.} & \textbf{Max Tok.} & \textbf{Benchmarks} & \textbf{Conditions} \\
\midrule
Claude 4 Sonnet & 0.0 & 8{,}192 & All 3 & All main experiments \\
Gemini 3.1 Pro & 0.0 & 8{,}192 & AutoDCW, MaTElDa & Debate, MaTElDa \\
Qwen3 235B & 0.0 & 8{,}192 & All 3 & Debate, cardinality, prompt sensitivity, MaTElDa, MMTU \\
DeepSeek R1 & 0.0 & 8{,}192 & All 3 & Debate, MaTElDa, MMTU \\
\bottomrule
\end{tabular}
\end{table}

\end{document}